\documentclass{article}
\PassOptionsToPackage{numbers,sort&compress}{natbib}
\usepackage[preprint]{neurips_2025}

\usepackage[utf8]{inputenc} 
\usepackage[T1]{fontenc}    
\usepackage{hyperref}       
\usepackage{url}            
\usepackage{booktabs}       
\usepackage{amsfonts}       
\usepackage{nicefrac}       
\usepackage{microtype}      
\usepackage{xcolor}         

\usepackage{amsmath}        
\usepackage{amsthm}         
\usepackage{amssymb}        
\usepackage{mathtools}
\usepackage{algorithm}      
\usepackage{algorithmic}    
\usepackage{multirow}       
\usepackage{wrapfig}        
\usepackage[capitalize]{cleveref}

\theoremstyle{plain}
\newtheorem{theorem}{Theorem}[section]
\newtheorem{proposition}[theorem]{Proposition}

\newtheorem{corollary}[theorem]{Corollary}
\theoremstyle{definition}

\theoremstyle{remark}

\newcommand{\E}{\mathbb{E}}

\DeclareMathOperator*{\argmin}{arg\,min}
\newcommand{\red}[1]{\textcolor{red}{#1}}

\title{Semi-gradient DICE for Offline Constrained Reinforcement Learning}

\author{
    Woosung Kim\textsuperscript{1*} \quad
    JunHo Seo\textsuperscript{1*} \quad
    Jongmin Lee\textsuperscript{2\dag} \quad
    Byung-Jun Lee\textsuperscript{1,3\dag} \\
    \\
    \textsuperscript{1}Korea University \quad
    \textsuperscript{2}Yonsei University \quad
    \textsuperscript{3}Gauss Labs Inc. \\
    \texttt{\{wsk208,junhoseo,byungjunlee\}@korea.ac.kr} \\
    \texttt{jongminlee@yonsei.ac.kr} \\
    \textsuperscript{*}Equal contribution \quad
    \textsuperscript{\dag}Corresponding authors
}

\begin{document}

\maketitle

\begin{abstract}
Stationary Distribution Correction Estimation (DICE) addresses the mismatch between the stationary distribution induced by a policy and the target distribution required for reliable off-policy evaluation (OPE) and policy optimization. DICE-based offline constrained RL particularly benefits from the flexibility of DICE, as it simultaneously maximizes return while estimating costs in offline settings. However, we have observed that recent approaches designed to enhance the offline RL performance of the DICE framework inadvertently undermine its ability to perform OPE, making them unsuitable for constrained RL scenarios. In this paper, we identify the root cause of this limitation: their reliance on a semi-gradient optimization, which solves a fundamentally different optimization problem and results in failures in cost estimation. Building on these insights, we propose a novel method to enable OPE and constrained RL through semi-gradient DICE. Our method ensures accurate cost estimation and achieves state-of-the-art performance on the offline constrained RL benchmark, DSRL.
\end{abstract}

\section{Introduction}

Constrained reinforcement learning (RL) trains agents to maximize return while adhering to predefined cost constraints. Conventional online RL involves interactions with the environment that can be unsafe or costly due to potential constraint violations in sensitive environment. To avoid the risk of violations during interactions, offline constrained RL offers a practical alternative by relying on a fixed dataset of pre-collected experiences, eliminating the need for potentially unsafe online exploration during training.

Specifically, offline constrained RL aims to maximize the expected return of a policy subject to cost constraints using only offline data. This objective makes stationary distribution correction estimation (DICE) a promising framework, leveraging stationary distribution to simultaneously estimate and optimize policy performance for both reward and cost functions \citep{polosky2022constrained, lee2021coptidice, zhangsafe}. Despite its theoretical appeal, prior DICE-based research has largely been limited to finite domains or struggled to achieve competitive performance in continuous state and action spaces compared to alternative frameworks.

Fortunately, recent empirical findings in offline RL \citep{sikchi2023dual, Mao2024odice} show that incorporating semi-gradient updates into the DICE objective significantly improves training stability and achieves state-of-the-art performance in large and continuous domains. However, our analysis reveals that applying semi-gradient methods causes DICE to lose its off-policy e valuation (OPE) capability, rendering them unsuitable for constrained scenarios.

Despite being were adopted to stabilize conflicting gradients, inspired by successful bootstrapped deep RL methods, we discovered that semi-gradient DICE algorithms inherently solve a fundamentally different optimization problem. This divergence yields a solution with distinct characteristics. We show that semi-gradient DICE algorithms are closely related to behavior-regularized offline RL, returning a \emph{policy correction} rather than the intended \emph{stationary distribution correction}. This observation partially explains the empirical success of the semi-gradient updates (\cref{Section4-analysis}).

Building on these analyses, we propose CORSDICE, an offline constrained RL algorithm. CORSDICE recovers a valid stationary distribution from the policy correction found by semi-gradient DICE, thereby enabling OPE while preserving the strong empirical performance (\cref{Section5-extraction}). We provide extensive empirical results validating our method across offline RL, OPE, and offline constrained RL tasks (\cref{Section6-experiment}).

\section{Preliminary and Related Works}
\paragraph{Markov Decision Process (MDP)} We model the environment as an infinite-horizon discounted Markov Decision Process (MDP) \citep{sutton2018rl}, defined by the tuple $\mathcal{M}:=\langle S, A, T, r, \gamma\rangle$, where $S$ and $A$ are the state and action spaces, $T: S\times A\to\Delta(S)$ is the transition distribution, $r: S\times A \to \mathbb{R}$ is the reward function, and $\gamma \in [0, 1)$ is a discount factor. Additionally, we define $p_0: \Delta(S)$ as the initial state distribution and a policy $\pi:S\to\Delta(A)$ as an action distribution conditioned on the state.

Given a policy $\pi$, its stationary distribution $d_\pi(s,a):=(1-\gamma)\sum^\infty_{t=0}\gamma^t\textrm{Pr}(s_t=s,a_t=a|\pi)$ represents the discounted probability of visiting a state-action pair $(s,a)$ when following $\pi$. These distributions satisfy the single-step transposed Bellman recurrence: $d_\pi(s,a)=(1-\gamma)p_0(s)\pi(a\mid s)+\gamma\pi(a\mid s)(\mathcal{T}_*d)(s)$,
where $(\mathcal{T}_*d)(s):=\sum_{\bar{s},\bar{a}}T(s\mid \bar{s},\bar{a})d(\bar{s},\bar{a})$. The expected discounted sum of rewards (return), or value $\rho(\pi)$, is defined as $\rho(\pi):=(1-\gamma)\mathbb{E}_{\pi}\left[\sum^\infty_{t=0}\gamma^t r(s_t, a_t)\right]$ and can be equivalently evaluated using stationary distribution $\rho(\pi)=\mathbb{E}_{(s,a)\sim d_\pi}[r(s,a)]
$.

\paragraph{OptiDICE} OptIDICE \citep{lee2021optidice} addresses distribution shift in offline RL by optimizing for a stationary distribution $d$ that maximizes expected return while being regularized towards the dataset distribution $d_D$ via an $f$-divergence penalty:
\begin{subequations}
\label{eq:OptiDICE-whole}
\begin{align}
\max_{d\geq0}\;&\sum_{s,a}d(s,a)r(s,a) - \alpha D_{f}(d||d_D)\label{eq:optidice_primal} \\
    \text{s.t.}\;&\sum_a d(s,a)=(1-\gamma)p_0(s)+\gamma(\mathcal{T}_*d)(s), \ \forall s\label{eq:bellman_flow}
\end{align}
\end{subequations}
where $D_{f}(d||d_D):=\mathbb{E}_{(s,a)\sim d_D} \big[ f\big( \frac{d(s,a)}{d_D(s,a)} \big) \big]$ denotes $f$-divergence. The solution $d^*$ to this problem is the stationary distribution of the policy $\pi^*$ maximizing the objective under Bellman flow constraints~\eqref{eq:bellman_flow}. OptiDICE~\citep{lee2021optidice} is implemented by solving the Lagrangian dual of \cref{eq:OptiDICE-whole} with multiplier $\nu(s)$ for~\eqref{eq:bellman_flow}:
\begin{equation}
    \min_{\nu}\max_{w\geq0}\; \mathcal{L}(w,\nu):=(1-\gamma)\mathbb{E}_{s_0\sim p_0}[\nu(s_0)] + \mathbb{E}_{(s,a) \sim d_D}\left[-\alpha f(w(s,a)) + w(s,a)e_{\nu}(s,a)\right]
    \label{eq:optidice_maxmin}
\end{equation}
where $d$ is replaced with the stationary distribution correction $w(s,a):=\tfrac{d(s,a)}{d_D(s,a)}$ to accommodate the offline dataset and $e_{\nu}(s,a):=r(s,a)+\gamma\mathbb{E}_{s'}[\nu(s')]-\nu(s)$ (derivation in \cref{Appendix-A.1}). 
Substituting $w$ with its closed-form solution $w^*_\nu(s,a)=\max\big(0, (f')^{-1}\big( \tfrac{e_\nu(s,a)}{\alpha} \big)\big)$ gives:
\begin{equation}
    \label{eq:fulldice}
    \min_{\nu}\;(1-\gamma)\mathbb{E}_{s_0\sim p_0}[\nu(s_0)] + \alpha\mathbb{E}_{ d_D}\left[f^*_0\big( \tfrac{e_\nu(s,a)}{\alpha} \big)\right]
\end{equation}
where $f^*_0(y):=\max_{x\geq0}xy-f(x)$ is a convex conjugate of $f$ in $\mathbb{R}^+$. A notable benefit of optimizing for $w^*$ is its applicability to OPE for any reward function, expressed as $\hat{\rho}(\pi)=\mathbb{E}_{(s,a)\sim d_D}[w^*_{\nu^*}(s,a)r(s,a)]$.

After obtaining $\nu^{*}$, a policy $\pi^*$ inducing the stationary distribution $w^{*}_{\nu^*}(s,a)d_D(s,a)$ must be extracted. For continuous action spaces without an analytical solution, weighted behavior cloning is commonly adopted by minimizing:
\begin{align}
    \label{pi-loss}
    \min_{\pi_{\theta}} \; - \mathbb{E}_{(s,a)\sim d_D}[w^*_{\nu^*}(s,a)\log \pi_{\theta}(a|s)] 
\end{align}
Despite this elegant formulation, OptiDICE's empirical performance in large and continuous domains lag behind compared to other value-based offline RL algorithms \cite{Mao2024odice}.

\paragraph{Semi-gradient Optimization} To improve OptiDICE's performance, semi-gradient variants have been explored \citep{sikchi2023dual, Mao2024odice}. This approach is a natural extension, inspired by fitted Q-iteration \citep{ernst2005tree} and motivated by the resemblance of the the residual $e_\nu(s,a)$ in \cref{eq:fulldice} to the Bellman error \citep{sutton2018rl}.

Prior semi-gradient methods \citep{sikchi2023dual, Mao2024odice} typically involved three modifications: (1) omitting the gradient from the next state $\nu(s')$ in $e_\nu(s,a)$, (2) replacing the initial state distribution $p_0(s)$ with the dataset distribution $d_D$,
and (3) introducing a temperature hyperparameter $\beta$ (often removing $\alpha$). Modification (3) is primarily a design choice and not directly tied to the core semi-gradient process.

To isolate and simplify the analysis of semi-gradient optimization, we adopt a semi-gradient algorithm that incorporates only modifications (1) and (2), specifically by entirely omitting the gradient of $\nu(s')$ and replacing $p_0$ with $d_D$. Introducing a function approximator $Q$, the resulting objectives are:
\begin{align}
\min_{\nu}\;&\mathbb{E}_{(s,a)\sim d_D} \big[\nu(s) + \alpha f^*_0 \big( \tfrac{Q(s,a)-\nu(s)}{\alpha} \big) \big]\\
\min_{Q}\;&\mathbb{E}_{(s,a,s')\sim d_D}[(r(s,a) + \gamma\nu(s')-Q(s,a)^2]
\end{align}
This formulation aligns with the semi-gradient optimization of \cref{eq:fulldice} when $Q$ is an exact optimum. Moreover, adopting $Q$ removes the bias introduced by estimating the expectation within a convex function in \cref{eq:fulldice} using finite samples - a limitation causing bias in OptiDICE in stochastic environments \citep{lee2021optidice,kim2024relaxed}. 
We refer to this specific algorithm as \textbf{SemiDICE} throughout the paper (details in \cref{appx:semi_dice}).

Based on this semi-gradient approach to DICE, \citet{sikchi2023dual, Mao2024odice} achieved state-of-the-art performance. Similarly, our empirical results confirm that SemiDICE exhibits strong offline RL performance consistent with those previous studies. Independently, \citet{xu2025idrl} proposed a similar theoretical formulation for semi-graident DICE within the context of unconstrained offline RL\footnote{We had independently derived the theoretical results (Propositions \ref{prop:solution_of_semi_dice}, \ref{proposition4.2}, and \ref{proposition5.1}) in an earlier version of this work, originally submitted to ICML 2024, prior to the appearance of \citet{xu2025idrl} (ICLR 2025).}. Our work complements theirs by offering comprehensive empirical validation and specifically applying and extending this formulation to address constrained offline RL.

\section{Constrained RL with DICE}
\label{sec:Constrained_RL_with_DICE}
In this section, we revisit the extension of the OptiDICE algorithm to constrained RL problems and explore the feasibility of extending SemiDICE using a similar approach.

\paragraph{Constrained RL} The constrained RL~\citep{altman1999cmdp} aims to obtain a policy that maximizes an expected return while satisfying cost constraints defined by a cost function $c: S\times A\to \mathbb{R}$ and a cost threshold  $C_\textrm{lim} \in \mathbb{R}$. The objective can be formulated as:

{\small \vspace{-15pt} \begin{align}
\label{eq:constrained_rl}
\max_{\pi}\mathbb{E}_{\pi} \textstyle \Big[ \sum\limits^{\infty}_{t=0}\gamma^{t}r(s_t,a_t)\Big]\;\textrm{s.t.}\;\mathbb{E}_{\pi}\Big[\sum\limits^{\infty}_{t=0}\gamma^t c(s_t,a_t)\leq C_{\textrm{lim}}\Big]
\end{align}}

\paragraph{COptiDICE}
The convenience of OPE using stationary distributions naturally extends to estimating the discounted sum of costs. 
In \citet{lee2021coptidice}, the constrained extension of OptiDICE solves Eq.~\eqref{eq:OptiDICE-whole} with the inclusion of additional constraints: 
\begin{align}
\sum_{s,a}d(s,a)c(s,a) \leq (1-\gamma)C_\textrm{lim}=:\tilde{C}_\textrm{lim}.
\label{eq:cost_constraint}
\end{align}
To satisfy the constraint, COptiDICE formulates the Lagrangian dual by adopting Lagrangian multiplier $\lambda$ for the cost constraint Eq.~\eqref{eq:cost_constraint}:
\begin{align*}
\min_{\nu,\lambda \geq 0}&\max_{w\geq0}\;\mathcal{L}(w,\nu) - \lambda \mathbb{E}_{d_D} [w(s,a)c(s,a) - \tilde{C}_\textrm{lim}])
\end{align*}
where $d$ is replaced with $w$ as in \cref{eq:optidice_maxmin}. Similar to OptiDICE, we can solve for the closed-form solution of $w$ and substitute it in to get training objective of $\nu$:
\begin{align*}
\min_{\nu}\;&(1-\gamma)\mathbb{E}_{s_0\sim p_0}[\nu(s_0)] + \alpha\mathbb{E}_{ d_D}\big[f^*_0\big( \tfrac{e_{\nu,\lambda}(s,a)}{\alpha}\big)\big]\\\min_{\lambda\geq0}\;&\lambda \left(\tilde{C}_\textrm{lim} - \mathbb{E}_{(s,a)\sim d_{D}}[w^{*}_{\nu, \lambda}(s,a)c(s,a)]\right)\\
& w^{*}_{\nu, \lambda}(s,a) = \max\big(0,\left(f'\right)^{-1}\big( \tfrac{e_{\nu,\lambda}(s,a)}{\alpha} \big)\big)
\end{align*}
where $e_{\nu,\lambda}(s,a):=r(s,a)-\lambda c(s,a)+\gamma\mathbb{E}_{s'}[\nu(s')]-\nu(s)$ (derivation in Appendix A.2). Looking at how $e_{\nu,\lambda}$ differs from previous $e_{\nu}$, we can interpret this algorithm as solving OptiDICE with a penalized reward function, $r(s,a)-\lambda c(s,a)$, where $\lambda$ is adjusted based on the cost constraint: it increases when the constraint is violated and decreases otherwise.

\paragraph{Constrained SemiDICE} 
As COptiDICE naturally extends OptiDICE to constrained RL, it initially appears feasible to extend SemiDICE in a similar manner to formulate a constrained SemiDICE, potentially enhancing performance in constrained RL problems. Naively applying the modifications introduced in SemiDICE to COptiDICE results in the following:

{\small \vspace{-15pt} \begin{align}
\min_{\nu}\; &\mathbb{E}_{d_D}\big[\nu(s)+\alpha f^*_0( \tfrac{Q(s,a) - \nu(s)}{\alpha})\big]\label{nu-loss}\\
\min_{Q}\; &\mathbb{E}_{d_D}\left[(r(s,a)-\lambda c(s,a)+\gamma\nu(s')-Q(s,a))^2\right]\label{Q-loss}\\
\min_{\lambda\geq0}\; & \lambda \left(\tilde{C}_\textrm{lim} - \mathbb{E}_{(s,a)\sim d_{D}}[w^{*}_{\nu, \lambda}(s,a)c(s,a)]\right) \label{pre-lambda-loss}
\end{align}}%
where $w^{*}_{\nu, \lambda}(s,a)=\max\big(0,(f')^{-1}\big(\frac{Q(s,a)-\nu(s)}{\alpha}\big)\big)$.

However, as will be described in ~\cref{Section4-analysis} and ~\cref{tab:exp_rmse}, this naive constrained SemiDICE completely fails to satisfy the cost constraint due to its inability to perform OPE correctly: SemiDICE estimates policy corrections rather than stationary distribution corrections, making $\mathbb{E}_{d_{D}}[w^{*}_{\nu, \lambda}(s,a)c(s,a)]$ no longer a valid cost estimate. 

\section{Demystifying SemiDICE}
\label{Section4-analysis}
In this section, we discuss various characteristics of the SemiDICE algorithm that deepen our understanding of semi-gradient optimization within the DICE framework. 

\paragraph{Solution of SemiDICE} We show that the correction $w(s,a)$ obtained by solving SemiDICE is not a stationary distribution correction, but rather a policy correction.

\begin{proposition}[Solution characteristics of SemiDICE]
\label{prop:solution_of_semi_dice}
 The correction $w^{*}(s,a)$ obtained by the optimal $\nu^*=\arg\min_\nu \mathbb{E}_{d_D}\big[\nu(s)+\alpha f^*_0(\tfrac{Q(s,a) - \nu(s)}{\alpha})\big]$, 
\begin{equation}
w^{*}(s,a)=\max\big(0,(f')^{-1}\big( \tfrac{Q(s,a)-\nu^{*}(s)}{\alpha} \big)\big),
\label{eq:semidice_w}
\end{equation} violates the Bellman flow constraint \cref{eq:bellman_flow} but satisfies the following conditions for $w^{*}(s,a)$ to act as a policy correction (Proof in \cref{appx:semi_dice_derivation}):
\begin{align}
\label{eq:policy-constraint}
\sum_a w^{*}(s,a)\pi_D(a|s) = 1, \ w^{*}(s,a) \geq 0, \ \forall s,a.
\end{align}
\end{proposition}
\cref{prop:solution_of_semi_dice} explains the failure of the naive constrained SemiDICE as $w(s,a)$ no longer converges to a stationary distribution correction under semi-gradient optimization, and resulting policy correction is incapable of performing OPE. The semi-gradient update has caused $\nu$ to lose its role as Lagrangian multiplier that ensures the satisfaction of Bellman flow constraints \eqref{eq:bellman_flow}. Similarly, other DICE algorithms employing semi-gradient optimization also fail to converge to stationary distribution corrections, instead converging to constant multiples of policy corrections~\citep{sikchi2023dual} or somewhere in-between two corrections~\citep{Mao2024odice} (\cref{appx:semi_dice_derivation,Appendix-B.2}). While not immediately obvious from \cref{prop:solution_of_semi_dice}, an important insight is that the Bellman flow violation is directly attributable to semi-gradient optimization, and is not caused by techniques such as the intital state replacement trick (refer to \cref{appx:semi_dice_derivation} for details regarding the trick).

For policy extraction, since weighted behavior cloning (\cref{pi-loss}) minimizes KL divergence, $\mathrm{KL}\big(\tfrac{w(s,a)d_D(s,a)}{\sum_a w(s,a)d_D(s,a)}\big\Vert \pi_\theta(a|s)\big)$, it remains effective regardless of the solution characteristics and has been successfully applied in previous studies.
As $w^*(s,a)$ computed by SemiDICE (Eq.~\eqref{eq:semidice_w}) represents the policy correction, we will denote it as $w(a|s)$ in later sections.

\paragraph{Connections to behavior-regularized offline RL} 
Having established that SemiDICE yields a policy correction, we now clarify the specific optimization problem it implicitly solves. Building upon the findings of prior works \citep{xuoffline,sikchi2023dual}, we demonstrate that SemiDICE, along with SQL \citep{xuoffline} and XQL \citep{garg2023extreme}, can be understood as solving behavior-regularized MDPs \citep{xuoffline,geist2019theory} with varying approximations.

Specifically, We consider a behavior-regularized MDP where the standard reward is augmented with penalty based on the reversed $f$-divergence between the dataset policy $\pi_D$ and learned policy $\pi$:
\begin{align*}
\max_{\pi} \ & \textstyle \E_{\pi}\Big[ \sum\limits_{t=0}^{\infty}\gamma^{t}\big(r(s_t,a_t)-\alpha \tfrac{\pi_D(a_t|s_t)}{\pi(a_t|s_t)}f\big( \tfrac{\pi(a_t|s_t)}{\pi_D(a_t|s_t)}\big)\big) \Big].
\end{align*}
This formulation still qualifies as a behavior-regularized MDP, as $xf(1/x)$ satisfies the necessary conditions for an $f$-divergence. The policy evaluation operator for this regularized MDP is given by:
\begin{align*}
(\mathcal{T}^{\pi}_{f} Q)(s,a) := r(s,a) + \gamma \mathbb{E}_{s' \sim T(\cdot|s,a)}[V(s')]
\end{align*}
where $V(s) :=\mathbb{E}_{a\sim\pi}\left[Q(s,a)-\alpha \frac{\pi_D(a|s)}{\pi(a|s)}f\left(\frac{\pi(a|s)}{\pi_D(a|s)}\right)\right]$.
\begin{proposition}
\label{proposition4.2}
In the behavior-regularized MDP, the optimal value functions $V^{*}(s), Q^{*}(s,a)$ and the optimal policy correction $\tfrac{\pi^{*}(a|s)}{\pi_D(a|s)}$ of the regularized MDP are given by (Proof in \cref{appendix-c.2-behavopt}):
\begin{align*}
    U^*(s) &= \argmin_{U(s)} U(s)+ \mathbb{E}_{a\sim\pi_D}\big[\alpha f^{*}(\tfrac{Q^{*}(s,a)-U(s)}{\alpha})\big] \\
    V^{*}(s)&= U^{*}(s)+\mathbb{E}_{a\sim\pi_D}\big[\alpha f^{*}(\tfrac{Q^{*}(s,a)-U^{*}(s)}{\alpha}) \big] \\
    Q^{*}(s,a)&= r(s,a) + \gamma \mathbb{E}_{s'\sim T}[V^{*}(s')] \\
    \tfrac{\pi^{*}(a|s)}{\pi_D(a|s)} &= \max\big(0,(f')^{-1}\big(\tfrac{Q^{*}(s,a)-U^{*}(s)}{\alpha}\big)\big)
\end{align*}
\end{proposition}

From \cref{proposition4.2}, we observe that SemiDICE is equivalent to approximating $V^{*}(s)$ with $U^{*}(s)$ in this behavior-regularized MDP. SQL makes a similar approximation ($V^{*}(s)$ with $U^{*}(s)+\alpha$) specifically for the $f(x) = x^2 - x$) case, while XQL corresponds to the $f(x) = x \log x$ case without approximation in $V^*(s)$ (details in \cref{appendix-c}).

The similarity between SemiDICE and behavior-regularized RL was also pointed out by \citet{Mao2024odice}. Yet, their investigation went as far as showing the equivalence of the optimization problem when employing \emph{KL divergence}. Our analysis, on the other hand, is applicable to general $f$-divergences and more distinctly reveals the distinctions and commonalities between these methods.

\paragraph{Advantage of semi-gradient update} 
Beyond the update stabilization previously identified by \citep{sikchi2023dual, Mao2024odice}, our analysis of SemiDICE's characteristics provides an alternative explanation for its superior performance over OptiDICE in large, continuous domains.

A key issue with OptiDICE is that the policy optimization can lead to the shrinking support of the policy and the states it visits. Depending on $f$-divergence, this can result in $d_{\pi^*}(s,a)=0$ for all $a$ in certain state $s$, even if $s$ is in the dataset. This state distribution sparsity problem worsens in larger state spaces, where the optimal policy $\pi^*(a|s)$ becomes undefined.

On the other hand, even with $f$-divergence that induce sparse solutions, semi-gradient DICE methods yield a sparse optimal \textbf{policy} (similar to \citealt{xuoffline}) rather than a sparse optimal \textbf{state-action stationary distribution}. This allows them to avoid the state distribution sparsity issue. 

When $d_{\pi^*}(s,a)=w^*(s,a) \red{d_D(s,a)} =0$, previously proposed DICE algorithms either resort to a uniform random policy (for finite action space; \citealt{zhan2022offline,ozdaglar2023revisiting,zhangsafe}) or refrain from updating the policy for the state $s$ (see \cref{pi-loss}), resulting in data inefficiency. In contrast, SemiDICE, by outputting policy corrections, inherently avoids this sparsity issue and the resulting data inefficiency.

\begin{corollary}
[SemiDICE avoiding sparsity problem]
\label{corollary2} Let $w^*$ be the correction (Eq.~\eqref{eq:semidice_w}) optimized by running SemiDICE. There is no state $s$ where $w^{*}(s,a) = 0 \ \forall a$. (Proof in \cref{appendix-c.3})
\end{corollary}

\section{CORSDICE: semi-gradient DICE for offline constrained RL}
\label{Section5-extraction}
Building on the finding that \emph{SemiDICE yields a policy correction}, we address the off-policy cost evaluation required for constrained SemiDICE (\cref{pre-lambda-loss}). This requires computing the state staionary distribution correction $w(s):=\frac{d_w(s)}{d_D(s)}$ from the policy correction $w(a|s) =\frac{\pi_w(a|s)}{\pi_D(a|s)}$ obtained via SemiDICE (\cref{eq:semidice_w}), where $d_w(s)$ is the state stationary distribution induced by the learned policy $\pi_w$. Successfully computing $w(s)$ enables off-policy cost evaluation using $w(s,a)=w(s)w(a|s)$, allowing optimization of $\lambda$ via:
\begin{align}
\min_{\lambda\geq0}\; \lambda \big(\tilde{C}_\textrm{lim} - \underbrace{\E_{(s,a)\sim d_{D}}[w(s)w(a|s)c(s,a)]}_{ = \E_{(s,a) \sim d^{\pi}} [ c(s,a) ]} \big). \label{modified-lambda-loss}
\end{align}
To compute $w(s)$ given $w(a|s)$, we introduce the following optimization problem:
\begin{subequations}
\label{eq:DICE_extraction}
\begin{align}
    &\max_{w(s)\geq 0}\;-\sum_{s}d_D(s)f\left(w(s)\right)\label{eq:extractobj}\\
    &\text{s.t.}\; w(s)d_D(s) = (1-\gamma)p_0(s) + \gamma(\mathcal{T}_{*}d_w)(s)\;\forall s \label{eq:extractconst}
\end{align}
\end{subequations}
where $(\mathcal{T}_*d_w)(s):=\sum_{\bar{s},\bar{a}}T(s\mid \bar{s},\bar{a})w(\bar{s})w(\bar{a}|\bar{s})d_D(\bar{s},\bar{a})$. The $|S|$ constraints \eqref{eq:extractconst} uniquely determine $w(s)$, making the problem over-constrained. However,  $f$-divergence objective \eqref{eq:extractobj} adds convexity, facilitating efficient sample-based optimization akin to dual of Q-LP methods \citep{nachum2020reinforcement}.
 
The Lagrangian dual problem, with multipliers $\mu(s)$ for constraint \eqref{eq:extractconst}, is given by:
\begin{equation}
\max_{w(s)\geq0}\min_{\mu(s)}\;(1-\gamma)\mathbb{E}_{s_0 \sim p_0}\left[\mu(s_0)\right] + \mathbb{E}_{(s,a)\sim d_D} \left[w(s)w(a|s)e_{\mu}(s,a)-f\left(w(s)\right)\right]\label{eq:sde_lag}
\end{equation}
where $e_{\mu}(s,a) = \gamma \sum_{s'}T(s'|s,a)\mu(s') - \mu(s)$ (Full derivation in \cref{Appendix-D}). Reversing the optimization order based on strong duality yields the a closed-form solution for $w^{*}(s)$:
\begin{align}
    w^{*}(s) \ &= \max(0,(f')^{-1} (\mathbb{E}_{a\sim \pi_D}[w(a|s)e_{\mu}(s,a)])).
\end{align}
Substituting $w^{*}(s)$ into \cref{eq:sde_lag} results in the dual objective for $\mu$:
\begin{equation}
\min_{\mu}\;\mathcal{L}_{\text{ext}}(\mu) := (1-\gamma)\mathbb{E}_{s_0 \sim p_0}\left[\mu(s_0)\right] + \mathbb{E}_{s\sim d_D} \left[f^{*}_{0}(\mathbb{E}_{a\sim \pi_D}[w(a|s)e_{\mu}(s,a)])\right].
\label{eq:extraction-0}
\end{equation}
Sample-based optimization of $\mathcal{L}_{\text{ext}}$ is challenging because expectations over $T$ and $\pi_D$ appears inside the convex function $f^{*}_{0}(x)$. Unlike previous algorithms biased only by transition stochasticity \citep{lee2021optidice,lee2021coptidice}, sampling actions from $\pi_D$ introduces significant additional bias when these expectations are estimated using a single sample.

\paragraph{Bias reduction for sample-based optimization}

Inspired by \citet{kim2024relaxed}, we propose a simple bias reduction technique by introducing an auxiliary function approximator $A(s)$ to estimate the expectation inside $f^*_0(\cdot)$ in \cref{eq:extraction-0}. This decomposes the $\mu$ optimization into the following optimizations for $A$ and $\mu$:
\begin{subequations}
\label{u-mu-loss}
\begin{align}
    \min_{A} \; &\mathbb{E}_{(s,a,s')\sim d_D}\left[ \big( A(s) - w(a|s)\hat{e}_{\mu}(s,s')\big)^2 \right]\label{u-loss}\\
    \min_{\mu} \; &\tilde{\mathcal{L}}_{\text{ext}}(\mu):=(1-\gamma)\mathbb{E}_{s_0 \sim p_0}\left[\mu(s_0)\right] \label{mu-loss} + \mathbb{E}_{(s,a,s')\sim d_D}\left[(f^*_0)'(A(s))w(a|s)\hat{e}_{\mu}(s,s') \right]
\end{align}
\end{subequations}
where $\hat{e}_{\mu}(s,s') = \gamma \mu(s')-\mu(s)$. We prove that minimizing these objectives results in the same optimal $\mu^{*}$ as in \cref{eq:extraction-0} (Proof in \cref{Appendix-D}).
\begin{proposition}\label{proposition5.1}
Minimization of the objectives in \eqref{u-mu-loss} results in the same optimal $\mu^{*}$ as in \eqref{eq:extraction-0}. 
\end{proposition}
While inspired by \citet{kim2024relaxed}, our technique addresses different bias sources: their method primarily tackles transition bias, whereas ours handles both transition and policy bias.

By optimizing the coupled objectives in \cref{u-loss,mu-loss}, we obtain the optimal $\mu^{*}$. Subsequently, the stationary distribution correction $w(s)$ is computed as $w(s)=\max(0,(f')^{-1}(A^{*}(s)))$. This provides necessary stationary distribution correction $w(s,a)=w(s)w(a|s)$ for accurate off-policy cost evaluation of the policy optimized by SemiDICE, as required for \cref{modified-lambda-loss}. While approximation errors may introduce some additional bias, our empirical observations suggest this bias is significantly smaller than that introduced by relying on the naive single-sample estimator for evaluating $\mathcal{L}_{\text{ext}}(\mu)$.

Leveraging these results, we propose \textit{\textbf{C}onstrained \textbf{O}ffline \textbf{R}L via \textbf{S}emi-gradient stationary \textbf{DI}stribution \textbf{C}orrection \textbf{E}stimation} (CORSDICE), an algorithm that iteratively alternates between optimizing the policy correction using SemiDICE on the penalized reward $r(s,a)-\lambda c(s,a)$ and performing off-policy cost evaluation via stationary distribution extraction, based on which $\lambda$ is updated. Algorithm~\ref{alg:algorithm} provides the complete procedure.

\section{Experiment}
\label{Section6-experiment}

\subsection{Examining algorithm characteristics}
\label{sec:experiment_on_random_finite_mdp}

\begin{figure*}[t]
\centering
\includegraphics[width=0.95\linewidth]{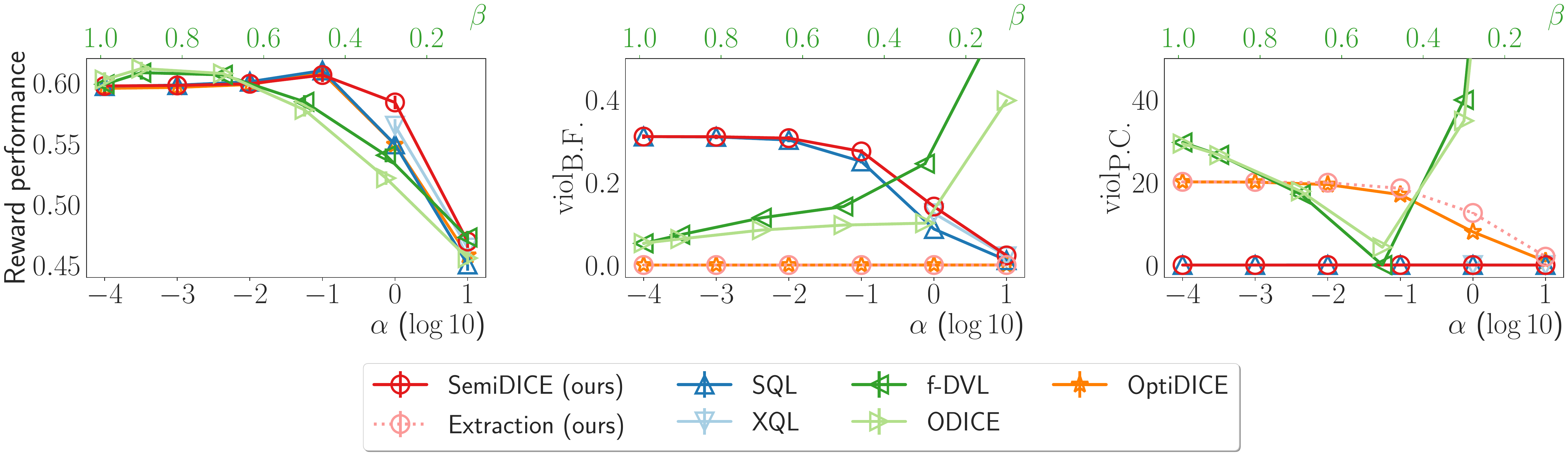}
\vspace{-5pt}
\caption{Policy Return (\textbf{Left}), Bellman flow constraint violation (\textbf{Middle}), and policy correction constraint violation (\textbf{Right}), averaged over $300$ runs. Hyperparameters $\alpha$ (SemiDICE, SQL, XQL, OptiDICE) and $\beta$ ($f$-DVL and ODICE) control $f$-divergence regularization strength (increasing with $\alpha$, decreasing with $\beta$). Performance of \textbf{XQL} for small $\alpha$ values is omitted due to numerical stability.}
\label{fig:random_mdp}
\vspace{-7pt}
\end{figure*}

We evaluate offline RL algorithms on a randomly generated finite MDP experiment~\citep{laroche2019safe, lee2020batch} (details in \cref{Appendix-F-Finite}) to empirically validate our finding that SemiDICE produces a policy correction and to demonstrate the effectiveness of our extraction method in recovering a valid stationary distribution. We compare four DICE-based methods (\textbf{OptiDICE}, \textbf{SemiDICE}, \textbf{f-DVL}, \textbf{ODICE}), two behvaior-regularized RL algorithms (\textbf{SQL}, \textbf{XQL}), and \textbf{Extraction}, an application of the state stationary distribution extraction method to \textbf{SemiDICE}. 

We analyze performance across three key aspects, visualized in \cref{fig:random_mdp}: \textbf{left}, policy performance $\rho(\pi)$; \textbf{middle}, violation of the Bellman flow constraint ($\text{viol}_{\text{B.F.}}$); and \textbf{right}, violation of the policy correction constraint ($\text{viol}_{\text{P.C.}}$). Violations are quantified using the $L_1$-norm:
\begin{align*}
    \text{viol}_{\text{B.F.}} &:= \sum_s \left\lvert(1-\gamma)p_0(s)+ \gamma(\mathcal{T}_{*}d_w)(s) - (\mathcal{B}_{*}d_w)(s)\right\rvert, \\
    \text{viol}_{\text{P.C.}} &:= \sum_s \Big\lvert \sum_a w(s,a)\pi_D(a|s)-1 \Big\rvert.
\end{align*}
The reward performance of all offline RL algorithms generally improves with $\alpha$ increases (decreasing $\beta$), but declines for large $\alpha$ (small $\beta$) due to strong conservatism shifting the policy toward suboptimal datasets. In tabular domains, when hyperparameters are properly tuned, performance differences among full-gradient (OptiDICE), semi-gradient (SemiDICE, f-DVL), and orthogonal-gradient (ODICE) are not significant. However, the benefits of semi-gradient optimization over full-gradient become more pronounced in large and continuous domain experiments resulting in significant performance differences (see \cref{Appendix-F-continuous}).

\paragraph{Solution characteristics}
The experiment validates whether semi-gradient DICE methods produce policy corrections rather than stationary distribution corrections, as theorized in \cref{prop:solution_of_semi_dice}.As shown in \cref{fig:random_mdp}, only \textbf{OptiDICE} and \textbf{Extraction} satisfy the Bellman-flow constraint (middle panel, zero violation). In contrast, \textbf{SemiDICE}, \textbf{SQL}, \textbf{XQL}, and \textbf{f-DVL ($\beta=0.5$)} yields policy corrections instead of stationary distribution corrections (right panel). These empirical observations align with our theoretical findings: SemiDICE outputs policy corrections, related to behavior-regularized RL algorithms, and our extraction method successfully recovers stationary distribution corrections from SemiDICE's result.

\subsection{Experiments on Off-Policy Evaluation}
\label{sec:experiment_on_off_policy_evaluation}

In this experiment, we estimate the returns of the SemiDICE policies on three D4RL~\citep{fu2020d4rl} benchmarks.\cref{tab:exp_rmse} presents root mean square error (RMSE) between estimated and average discounted returns. We compare four algorithms: \textbf{SemiDICE}, which directly uses the policy correction $w(a|s)$ for OPE ($\rho(\pi_w)=\mathbb{E}_{d_D}[w(a|s)r(s,a)]$); \textbf{Extraction}, which uses the extracted stationary distribution correction ($\rho(\pi_w)=\mathbb{E}_{d_D}[w(s)w(a|s)r(s,a)]$); \textbf{IHOPE} \citep{liu2018breaking}, which extracts a stationary distribution from the policy correction using an adversarial discriminator; and \textbf{DualDICE} \citep{nachum2019dualdice}, a representative Q-LP-based OPE method.\footnote{Our policy correction-based approach slightly deviates from conventional OPE settings \citep{nachum2019dualdice, yang2020off}, which typically assume direct policy access. Applying convention OPE algorithms thus requires prior policy extraction.}

\begin{wraptable}{r}{0.4\textwidth}
\vspace{-15pt}
\caption{RMSE of OPE for SemiDICE policies trained on D4RL~\cite{fu2020d4rl} dataset (more results in \cref{sec:appx_exp_d4rl}).}
\label{tab:exp_rmse}
\centering
\resizebox{0.4\textwidth}{!}{
\begin{tabular}{lccc}
\toprule
Algo. & Hopper & HalfCheetah & Walker2d \\
\midrule
SemiDICE & 90.62 & 87.58 & 111.63 \\
Ext. (Ours) & \textbf{20.70} & \textbf{26.44} & \textbf{9.20} \\
DualDICE & 58.08 & 162.81 & 20.53 \\
IHOPE & 78.61 & 57.92 & 90.01 \\
\bottomrule
\end{tabular}
}
\vspace{-10pt}
\end{wraptable}

\cref{tab:exp_rmse} shows that \textbf{SemiDICE} consistently underperforms, confirming that using the policy correction alone is unsuitable for OPE. In contrast, $\textbf{Extraction}$ successfully derives a valid stationary distribution, outperforming the baselines. Compared to the extensively studied Q-LP methods like \textbf{DualDICE}, our extraction benefits from in-sample learning, avoiding evaluation of OOD actions by the learned function approximator. While $\textbf{IHOPE}$ also computes the marginalized correction from the policy correction, its reliance on min-max optimization reduces training stability. Our algorithm, based on single convex optimization, ensures greater stability and superior estimation performance.

\begin{table*}[t]
\vspace{-10pt}
\caption{Normalized DSRL~\cite{liu2024offlinesaferl} benchmark results averaged over 5 seeds and 20 runs. {\color[HTML]{656565} \textbf{Gray}}: Unsafe agents, \textbf{Bold}: Safe agents whose normalized costs below 1.0, {\color[HTML]{0000FF} \textbf{Blue}}: Safe agents achieving the highest normalized return.}
\vspace{-10pt}
\label{tab:exp_dsrl}
\begin{center}
\begin{small}
\begin{sc}
\resizebox{1.\linewidth}{!}{
\begin{tabular}{c cc cc cc cc cc cc cc}
\toprule
\multirow{2}{*}{Task} & \multicolumn{2}{c}{BC-All} & \multicolumn{2}{c}{BC-Safe} & \multicolumn{2}{c}{BCQ-Lag} & \multicolumn{2}{c}{BEAR-Lag} & \multicolumn{2}{c}{CPQ} & \multicolumn{2}{c}{COptiDICE} & \multicolumn{2}{c}{CORSDICE (ours)} \\
\cmidrule(r){2-3} \cmidrule(r){4-5} \cmidrule(r){6-7} \cmidrule(r){8-9} \cmidrule(r){10-11} \cmidrule(r){12-13} \cmidrule(r){14-15}
& reward $\uparrow$ & cost $\downarrow$ & reward $\uparrow$ & cost $\downarrow$ & reward $\uparrow$ & cost $\downarrow$ & reward $\uparrow$ & cost $\downarrow$ & reward $\uparrow$ & cost $\downarrow$ & reward $\uparrow$ & cost $\downarrow$ & reward $\uparrow$ & cost $\downarrow$ \\
\midrule
PointButton1 & {\color[HTML]{656565} 0.14} & {\color[HTML]{656565} 1.01} & \textbf{0.06} & \textbf{0.60} & {\color[HTML]{656565} 0.38} & {\color[HTML]{656565} 2.69} & {\color[HTML]{656565} 0.60} & {\color[HTML]{656565} 3.47} & {\color[HTML]{656565} 0.70} & {\color[HTML]{656565} 4.07} & \textbf{0.15} & \textbf{1.00} & \textbf{\color[HTML]{0000FF} 0.22} & \textbf{\color[HTML]{0000FF} 0.94} \\
PointButton2 & {\color[HTML]{656565} 0.26} & {\color[HTML]{656565} 1.62} & {\color[HTML]{656565} 0.16} & {\color[HTML]{656565} 1.02} & {\color[HTML]{656565} 0.45} & {\color[HTML]{656565} 2.78} & {\color[HTML]{656565} 0.65} & {\color[HTML]{656565} 3.63} & {\color[HTML]{656565} 0.64} & {\color[HTML]{656565} 3.30} & {\color[HTML]{656565} 0.26} & {\color[HTML]{656565} 1.61} & \textbf{\color[HTML]{0000FF} 0.13} & \textbf{\color[HTML]{0000FF} 0.98} \\
PointCircle1 & {\color[HTML]{656565} 0.78} & {\color[HTML]{656565} 4.80} & \textbf{0.41} & \textbf{0.20} & {\color[HTML]{656565} 0.83} & {\color[HTML]{656565} 4.65} & {\color[HTML]{656565} 0.34} & {\color[HTML]{656565} 2.31} & \textbf{\color[HTML]{0000FF} 0.54} & \textbf{\color[HTML]{0000FF} 0.29} & {\color[HTML]{656565} 0.80} & {\color[HTML]{656565} 4.00} & \textbf{0.43} & \textbf{0.93} \\
PointCircle2 & {\color[HTML]{656565} 0.67} & {\color[HTML]{656565} 4.89} & \textbf{0.47} & \textbf{0.96} & {\color[HTML]{656565} 0.65} & {\color[HTML]{656565} 3.77} & {\color[HTML]{656565} 0.26} & {\color[HTML]{656565} 3.84} & {\color[HTML]{656565} 0.32} & {\color[HTML]{656565} 1.18} & {\color[HTML]{656565} 0.64} & {\color[HTML]{656565} 4.18} & \textbf{\color[HTML]{0000FF} 0.49} & \textbf{\color[HTML]{0000FF} 0.75}\\
PointGoal1 & \textbf{0.64} & \textbf{0.93} & \textbf{0.42} & \textbf{0.35} & {\color[HTML]{656565} 0.72} & {\color[HTML]{656565} 1.02} & {\color[HTML]{656565} 0.77} & {\color[HTML]{656565} 1.18} & \textbf{0.44} & \textbf{0.62} & \textbf{0.63} & \textbf{0.96} & \textbf{\color[HTML]{0000FF} 0.75} & \textbf{\color[HTML]{0000FF} 0.89} \\
PointGoal2 & {\color[HTML]{656565} 0.53} & {\color[HTML]{656565} 2.00} & \textbf{0.27} & \textbf{0.76} & {\color[HTML]{656565} 0.74} & {\color[HTML]{656565} 3.72} & {\color[HTML]{656565} 0.84} & {\color[HTML]{656565} 3.94} & {\color[HTML]{656565} 0.50} & {\color[HTML]{656565} 1.26} & {\color[HTML]{656565} 0.55} & {\color[HTML]{656565} 2.08} & \textbf{\color[HTML]{0000FF} 0.37} & \textbf{\color[HTML]{0000FF} 0.85}\\
PointPush1 & \textbf{0.23} & \textbf{0.88} & \textbf{0.16} & \textbf{0.52} & {\color[HTML]{656565} 0.36} & {\color[HTML]{656565} 1.08} & {\color[HTML]{656565} 0.44} & {\color[HTML]{656565} 1.01} & {\color[HTML]{656565} 0.26} & {\color[HTML]{656565} 1.27} & \textbf{0.24} & \textbf{0.74} & \textbf{\color[HTML]{0000FF} 0.27} & \textbf{\color[HTML]{0000FF} 0.81} \\
PointPush2 & {\color[HTML]{656565} 0.14} & {\color[HTML]{656565} 1.21} & \textbf{0.12} & \textbf{0.59} & {\color[HTML]{656565} 0.25} & {\color[HTML]{656565} 1.51} & {\color[HTML]{656565} 0.27} & {\color[HTML]{656565} 1.81} & {\color[HTML]{656565} 0.14} & {\color[HTML]{656565} 1.55} & \textbf{0.15} & \textbf{0.86} & \textbf{\color[HTML]{0000FF} 0.18} & \textbf{\color[HTML]{0000FF} 0.07} \\
CarButton1 & {\color[HTML]{656565} 0.16} & {\color[HTML]{656565} 1.73} & \textbf{0.05} & \textbf{0.50} & {\color[HTML]{656565} 0.44} & {\color[HTML]{656565} 7.50} & {\color[HTML]{656565} 0.53} & {\color[HTML]{656565} 7.49} & {\color[HTML]{656565} 0.53} & {\color[HTML]{656565} 8.26} & {\color[HTML]{656565} 0.00} & {\color[HTML]{656565} 1.40} & \textbf{\color[HTML]{0000FF} 0.09} & \textbf{\color[HTML]{0000FF} 0.48} \\
CarButton2 & {\color[HTML]{656565} -0.13} & {\color[HTML]{656565} 1.78} & \textbf{0.03} & \textbf{0.67} & {\color[HTML]{656565} 0.53} & {\color[HTML]{656565} 6.12} & {\color[HTML]{656565} 0.60} & {\color[HTML]{656565} 6.24} & {\color[HTML]{656565} 0.61} & {\color[HTML]{656565} 5.03} & {\color[HTML]{656565} -0.04} & {\color[HTML]{656565} 1.23} & \textbf{\color[HTML]{0000FF} 0.06} & \textbf{\color[HTML]{0000FF} 0.71} \\
CarCircle1 & {\color[HTML]{656565} 0.72} & {\color[HTML]{656565} 5.32} & {\color[HTML]{656565} 0.30} & {\color[HTML]{656565} 1.32} & {\color[HTML]{656565} 0.76} & {\color[HTML]{656565} 4.95} & {\color[HTML]{656565} 0.81} & {\color[HTML]{656565} 6.78} & {\color[HTML]{656565} 0.03} & {\color[HTML]{656565} 2.41} & {\color[HTML]{656565} 0.71} & {\color[HTML]{656565} 4.91} & \textbf{\color[HTML]{0000FF} -0.05} & \textbf{\color[HTML]{0000FF} 0.64}\\
CarCircle2 & {\color[HTML]{656565} 0.69} & {\color[HTML]{656565} 6.42} & {\color[HTML]{656565} 0.40} & {\color[HTML]{656565} 2.19} & {\color[HTML]{656565} 0.69} & {\color[HTML]{656565} 6.18} & {\color[HTML]{656565} 0.83} & {\color[HTML]{656565} 10.45} & \textbf{\color[HTML]{0000FF} 0.52} & \textbf{\color[HTML]{0000FF} 0.41} & {\color[HTML]{656565} 0.68} & {\color[HTML]{656565} 6.00} & \textbf{0.33} & \textbf{0.78} \\
CarGoal1 & \textbf{0.40} & \textbf{0.54} & \textbf{0.29} & \textbf{0.39} & \textbf{0.50} & \textbf{0.95} & {\color[HTML]{656565} 0.71} & {\color[HTML]{656565} 1.29} & \textbf{\color[HTML]{0000FF} 0.81} & \textbf{\color[HTML]{0000FF} 0.94} & \textbf{0.51} & \textbf{0.82} & \textbf{0.53} & \textbf{0.79} \\
CarGoal2 & {\color[HTML]{656565} 0.28} & {\color[HTML]{656565} 1.06} & \textbf{0.16} & \textbf{0.49} & {\color[HTML]{656565} 0.69} & {\color[HTML]{656565} 3.51} & {\color[HTML]{656565} 0.83} & {\color[HTML]{656565} 3.74} & {\color[HTML]{656565} 0.88} & {\color[HTML]{656565} 4.26} & {\color[HTML]{656565} 0.33} & {\color[HTML]{656565} 1.24} & \textbf{\color[HTML]{0000FF} 0.39} & \textbf{\color[HTML]{0000FF} 0.99} \\
CarPush1 & \textbf{0.22} & \textbf{0.56} & \textbf{0.18} & \textbf{0.46} & \textbf{0.36} & \textbf{0.73} & \textbf{0.43} & \textbf{0.83} & {\color[HTML]{656565} 0.15} & {\color[HTML]{656565} 1.33} & \textbf{0.22} & \textbf{0.56} & \textbf{0.22} & \textbf{0.71} \\
CarPush2 & {\color[HTML]{656565} 0.12} & {\color[HTML]{656565} 1.49} & \textbf{0.05} & \textbf{0.40} & {\color[HTML]{656565} 0.38} & {\color[HTML]{656565} 2.68} & {\color[HTML]{656565} 0.35} & {\color[HTML]{656565} 2.78} & {\color[HTML]{656565} 0.29} & {\color[HTML]{656565} 3.62} & {\color[HTML]{656565} 0.13} & {\color[HTML]{656565} 1.15} & \textbf{\color[HTML]{0000FF} 0.15} & \textbf{\color[HTML]{0000FF} 0.91} \\
SwimmerVel & {\color[HTML]{656565} 0.47} & {\color[HTML]{656565} 1.05} & \textbf{0.47} & \textbf{0.31} & {\color[HTML]{656565} 0.28} & {\color[HTML]{656565} 2.35} & \textbf{0.17} & \textbf{0.84} & {\color[HTML]{656565} 0.59} & {\color[HTML]{656565} 3.18} & \textbf{\color[HTML]{0000FF} 0.56} & \textbf{\color[HTML]{0000FF} 0.64} & \textbf{0.09} & \textbf{0.38} \\
HopperVel & {\color[HTML]{656565} 0.85} & {\color[HTML]{656565} 3.78} & {\color[HTML]{656565} 0.61} & {\color[HTML]{656565} 2.11} & {\color[HTML]{656565} 0.46} & {\color[HTML]{656565} 2.18} & \textbf{-0.01} & \textbf{0.0} & {\color[HTML]{656565} 0.46} & {\color[HTML]{656565} 3.08}  & {\color[HTML]{656565} 0.86} & {\color[HTML]{656565} 1.60} & \textbf{\color[HTML]{0000FF} 0.80} & \textbf{\color[HTML]{0000FF} 0.41} \\
HalfCheetahVel & {\color[HTML]{656565} 0.89} & {\color[HTML]{656565} 2.57} & \textbf{0.88} & \textbf{0.13} & {\color[HTML]{656565} 0.90} & {\color[HTML]{656565} 1.16} & \textbf{0.03} & \textbf{0.01} & \textbf{0.32} & \textbf{0.95} & \textbf{\color[HTML]{0000FF} 0.96} & \textbf{\color[HTML]{0000FF} 0.24} & \textbf{0.95} & \textbf{0.30} \\
Walker2dVel & {\color[HTML]{656565} 0.81} & {\color[HTML]{656565} 1.19} & \textbf{0.79} & \textbf{0.00} & {\color[HTML]{656565} 0.64} & {\color[HTML]{656565} 2.53} & \textbf{0.01} & \textbf{0.00} & \textbf{0.00} & \textbf{0.02} & \textbf{\color[HTML]{0000FF} 0.80} & \textbf{\color[HTML]{0000FF} 0.87} & \textbf{\color[HTML]{0000FF} 0.80} & \textbf{\color[HTML]{0000FF} 0.02} \\
AntVel & {\color[HTML]{656565} 0.98} & {\color[HTML]{656565} 4.73} & \textbf{0.97} & \textbf{0.36} & {\color[HTML]{656565} 0.66} & {\color[HTML]{656565} 4.03} & \textbf{0.32} & \textbf{0.12} & \textbf{-0.14} & \textbf{0.04} & {\color[HTML]{656565} 1.01} & {\color[HTML]{656565} 1.24} & \textbf{\color[HTML]{0000FF} 0.98} & \textbf{\color[HTML]{0000FF} 0.23}\\
\midrule
\textbf{SafetyGym Average} & {\color[HTML]{656565} 0.47} & {\color[HTML]{656565} 2.36} & \textbf{0.35} & \textbf{0.68} & {\color[HTML]{656565} 0.56} & {\color[HTML]{656565} 3.15} & {\color[HTML]{656565} 0.47} & {\color[HTML]{656565} 2.94} & {\color[HTML]{656565} 0.41} & {\color[HTML]{656565} 2.19} & {\color[HTML]{656565} 0.48} & {\color[HTML]{656565} 1.78} & \textbf{\color[HTML]{0000FF} 0.40} & \textbf{\color[HTML]{0000FF} 0.71} \\
\midrule
BallRun & {\color[HTML]{656565} 0.43} & {\color[HTML]{656565} 1.10} & {\color[HTML]{656565} 0.25} & {\color[HTML]{656565} 1.15} & {\color[HTML]{656565} 0.96} & {\color[HTML]{656565} 2.49} & {\color[HTML]{656565} 0.02} & {\color[HTML]{656565} 1.56} & {\color[HTML]{656565} 0.64} & {\color[HTML]{656565} 2.70} & \textbf{-0.01} & \textbf{0.00} & \textbf{\color[HTML]{0000FF} 0.25} & \textbf{\color[HTML]{0000FF} 0.99} \\
CarRun & \textbf{\color[HTML]{0000FF} 0.97} & \textbf{\color[HTML]{0000FF} 0.15} & \textbf{\color[HTML]{0000FF} 0.97} & \textbf{\color[HTML]{0000FF} 0.12} & \textbf{0.96} & \textbf{0.32} & \textbf{-0.54} & \textbf{0.37} & \textbf{0.89} & \textbf{0.41} & \textbf{0.30} & \textbf{0.01} & \textbf{\color[HTML]{0000FF} 0.97} & \textbf{\color[HTML]{0000FF} 0.55} \\
DroneRun & {\color[HTML]{656565} 0.56} & {\color[HTML]{656565} 1.73} & {\color[HTML]{656565} 0.43} & {\color[HTML]{656565} 1.14} & {\color[HTML]{656565} 0.58} & {\color[HTML]{656565} 1.96} & {\color[HTML]{656565} -0.18} & {\color[HTML]{656565} 5.40} & {\color[HTML]{656565} 0.40} & {\color[HTML]{656565} 1.40} & {\color[HTML]{656565} 0.59} & {\color[HTML]{656565} 1.42} & \textbf{\color[HTML]{0000FF} 0.48} & \textbf{\color[HTML]{0000FF} 0.97} \\
AntRun & {\color[HTML]{656565} 0.73} & {\color[HTML]{656565} 1.55} & \textbf{\color[HTML]{0000FF} 0.69} & \textbf{\color[HTML]{0000FF} 0.95} & \textbf{0.63} & \textbf{0.92} & \textbf{0.27} & \textbf{0.25} & \textbf{0.19} & \textbf{0.43} & {\color[HTML]{656565} 0.73} & {\color[HTML]{656565} 1.35} & \textbf{0.66} & \textbf{0.51} \\
BallCircle & {\color[HTML]{656565} 0.72} & {\color[HTML]{656565} 1.13} & \textbf{0.40} & \textbf{0.55} & {\color[HTML]{656565} 0.87} & {\color[HTML]{656565} 1.52} & {\color[HTML]{656565} 0.31} & {\color[HTML]{656565} 1.54} & \textbf{\color[HTML]{0000FF} 0.74} & \textbf{\color[HTML]{0000FF} 0.75} & {\color[HTML]{656565} 0.78} & {\color[HTML]{656565} 1.18} & \textbf{0.56} & \textbf{0.64} \\
CarCircle & {\color[HTML]{656565} 0.72} & {\color[HTML]{656565} 1.11} & {\color[HTML]{656565} 0.18} & {\color[HTML]{656565} 1.11} & {\color[HTML]{656565} 0.65} & {\color[HTML]{656565} 2.48} & {\color[HTML]{656565} 0.15} & {\color[HTML]{656565} 2.5} & \textbf{\color[HTML]{0000FF} 0.70} & \textbf{\color[HTML]{0000FF} 0.66} & {\color[HTML]{656565} 0.76} & {\color[HTML]{656565} 1.28} & \textbf{0.34} & \textbf{0.64} \\
DroneCircle & {\color[HTML]{656565} 0.68} & {\color[HTML]{656565} 1.17} & \textbf{\color[HTML]{0000FF} 0.55} & \textbf{\color[HTML]{0000FF} 0.42} & \textbf{0.50} & \textbf{0.24} & \textbf{-0.11} & \textbf{0.35} & {\color[HTML]{656565} -0.11} & {\color[HTML]{656565} 1.31} & {\color[HTML]{656565} 0.84} & {\color[HTML]{656565} 1.16} & \textbf{\color[HTML]{0000FF} 0.55} & \textbf{\color[HTML]{0000FF} 0.77} \\
AntCircle & {\color[HTML]{656565} 0.71} & {\color[HTML]{656565} 2.83} & {\color[HTML]{656565} 0.48} & {\color[HTML]{656565} 1.26} & {\color[HTML]{656565} 0.84} & {\color[HTML]{656565} 4.29} & \textbf{0.22} & \textbf{0.52} & \textbf{0.00} & \textbf{0.00} & {\color[HTML]{656565} 0.74} & {\color[HTML]{656565} 5.39} & \textbf{\color[HTML]{0000FF} 0.53} & \textbf{\color[HTML]{0000FF} 0.88} \\
\midrule
\textbf{BulletGym Average} & {\color[HTML]{656565} 0.69} & {\color[HTML]{656565} 1.35} & \textbf{0.49} & \textbf{0.84} &  {\color[HTML]{656565} 0.75} & {\color[HTML]{656565} 1.78} & {\color[HTML]{656565} 0.02} & {\color[HTML]{656565} 1.56} & \textbf{0.43} & \textbf{0.96} & {\color[HTML]{656565} 0.59} & {\color[HTML]{656565} 1.47} & {\color[HTML]{0000FF} \textbf{0.54}} & {\color[HTML]{0000FF} \textbf{0.77}} \\
\midrule
easysparse & \textbf{0.27} & \textbf{0.32} & \textbf{0.32} & \textbf{0.06} & {\color[HTML]{656565} 1.37} & {\color[HTML]{656565}3.10} & \textbf{-0.03} & \textbf{0.05} & \textbf{-0.23} & \textbf{0.17} & {\color[HTML]{656565} 0.91} & {\color[HTML]{656565} 2.64} & \textbf{\color[HTML]{0000FF} 0.54} & \textbf{\color[HTML]{0000FF} 0.85} \\
easymean & {\color[HTML]{656565} 0.51} & {\color[HTML]{656565} 1.45} & \textbf{0.25} & \textbf{0.00} & {\color[HTML]{656565} 1.31} & {\color[HTML]{656565} 2.67} & \textbf{-0.03} & \textbf{0.07} & \textbf{-0.06} & \textbf{0.02} & {\color[HTML]{656565} 0.75} & {\color[HTML]{656565} 2.67} & \textbf{\color[HTML]{0000FF} 0.49} & \textbf{\color[HTML]{0000FF} 0.91}\\
easydense & {\color[HTML]{656565} 0.64} & {\color[HTML]{656565} 2.21} & \textbf{0.22} & \textbf{0.01} & {\color[HTML]{656565} 1.02} & {\color[HTML]{656565} 1.99} & \textbf{0.09} & \textbf{0.46} & \textbf{-0.06} & \textbf{0.02} & {\color[HTML]{656565} 0.70} & {\color[HTML]{656565} 1.13} & \textbf{\color[HTML]{0000FF} 0.52} & \textbf{\color[HTML]{0000FF} 0.94} \\
mediumsparse & {\color[HTML]{656565} 0.81} & {\color[HTML]{656565} 1.15} & \textbf{0.74} & \textbf{0.14} & \textbf{0.77} & \textbf{0.72} & \textbf{-0.03} & \textbf{0.02} & \textbf{-0.08} & \textbf{0.01} & {\color[HTML]{656565} 0.83} & {\color[HTML]{656565} 1.51} & \textbf{\color[HTML]{0000FF} 0.99} & \textbf{\color[HTML]{0000FF} 0.94}\\
mediummean & {\color[HTML]{656565} 0.77} & {\color[HTML]{656565} 1.37} & \textbf{0.72} & \textbf{0.25} & {\color[HTML]{656565} 2.03} & {\color[HTML]{656565} 2.60} & \textbf{-0.02} & \textbf{0.03} & \textbf{-0.08} & \textbf{0.02} & {\color[HTML]{656565} 0.92} & {\color[HTML]{656565} 1.89} & \textbf{\color[HTML]{0000FF} 0.96} & \textbf{\color[HTML]{0000FF} 0.76}\\
mediumdense & {\color[HTML]{656565} 0.81} & {\color[HTML]{656565} 1.26} & \textbf{0.82} & \textbf{0.82} & {\color[HTML]{656565} 2.20} & {\color[HTML]{656565} 2.79} & \textbf{0.06} & \textbf{0.16} & \textbf{-0.07} & \textbf{0.00} & \textbf{0.73} & \textbf{0.89} & \textbf{\color[HTML]{0000FF} 0.98} & \textbf{\color[HTML]{0000FF} 0.83} \\
hardsparse & {\color[HTML]{656565} 0.46} & {\color[HTML]{656565} 2.07} & \textbf{0.37} & \textbf{0.19} & {\color[HTML]{656565} 1.15} & {\color[HTML]{656565} 2.78} & \textbf{0.01} & \textbf{0.28} & \textbf{-0.04} & \textbf{0.01} & {\color[HTML]{656565} 0.56} & {\color[HTML]{656565} 1.64} & \textbf{\color[HTML]{0000FF} 0.41} & \textbf{\color[HTML]{0000FF} 0.84} \\
hardmean & {\color[HTML]{656565} 0.36} & {\color[HTML]{656565} 1.14} & \textbf{0.32} & \textbf{0.08} & {\color[HTML]{656565} 0.94} & {\color[HTML]{656565} 2.18} & \textbf{0.00} & \textbf{0.11} & \textbf{-0.05} & \textbf{0.01} & {\color[HTML]{656565} 0.64} & {\color[HTML]{656565} 1.14} & \textbf{\color[HTML]{0000FF} 0.42} & \textbf{\color[HTML]{0000FF} 0.83} \\
harddense & {\color[HTML]{656565} 0.40} & {\color[HTML]{656565} 1.70} & \textbf{0.29} & \textbf{0.10} & {\color[HTML]{656565} 1.19} & {\color[HTML]{656565} 3.00} & \textbf{0.00} & \textbf{0.05} & \textbf{-0.04} & \textbf{0.00} & \textbf{\color[HTML]{0000FF} 0.51} & \textbf{\color[HTML]{0000FF} 0.72} & \textbf{0.27} & \textbf{0.59}\\
\midrule
\textbf{MetaDrive Average} & {\color[HTML]{656565} 0.56} & {\color[HTML]{656565} 1.41} & \textbf{0.45} & \textbf{0.18} &  {\color[HTML]{656565} 1.33} & {\color[HTML]{656565} 2.43} & \textbf{0.01} & \textbf{0.14} & \textbf{-0.08} & \textbf{0.03} & {\color[HTML]{656565} 0.73} & {\color[HTML]{656565} 1.58} & {\color[HTML]{0000FF} \textbf{0.62}} & {\color[HTML]{0000FF} \textbf{0.85}} \\
\bottomrule
\end{tabular}
}
\end{sc}
\end{small}
\end{center}
\vspace{-12pt}
\end{table*}

\begin{table*}[t]
\vspace{-5pt}
\caption{Normalized DSRL~\citep{liu2024offlinesaferl} with advanced function approximators and tighter cost limits. The results of baselines with asterisk (*) are adopted from FISOR~\citep{zheng2024safe}. {\color[HTML]{656565}\textbf{Gray}}: Unsafe agents, \textbf{Bold}: Safe agents whose normalized costs are below 1.0, \textbf{\color[HTML]{0000FF} Blue}: Safe agents with the highest normalized return.}
\vspace{-10pt}
\label{tab:exp_adv_dsrl}
\begin{center}
\begin{small}
\begin{sc}
\resizebox{1.\linewidth}{!}{
\begin{tabular}{c cc cc cc cc cc cc}
\toprule
\multirow{2}{*}{Task} & \multicolumn{2}{c}{D-BC-All} & \multicolumn{2}{c}{D-BC-Safe} & \multicolumn{2}{c}{CDT*} & \multicolumn{2}{c}{TREBI*} & \multicolumn{2}{c}{FISOR*} & \multicolumn{2}{c}{D-CORSDICE (Ours)} \\
\cmidrule(r){2-3} \cmidrule(r){4-5} \cmidrule(r){6-7} \cmidrule(r){8-9} \cmidrule(r){10-11} \cmidrule(r){12-13} 
& reward $\uparrow$ & cost $\downarrow$ & reward $\uparrow$ & cost $\downarrow$ & reward $\uparrow$ & cost $\downarrow$ & reward $\uparrow$ & cost $\downarrow$ & reward $\uparrow$ & cost $\downarrow$ & reward $\uparrow$ & cost $\downarrow$ \\
\midrule
CarButton1 & {\color[HTML]{656565} 0.15} & {\color[HTML]{656565} 14.50} & {\color[HTML]{656565} 0.03} & {\color[HTML]{656565} 5.25} & {\color[HTML]{656565} 0.17} & {\color[HTML]{656565} 7.05} & {\color[HTML]{656565} 0.07} & {\color[HTML]{656565} 3.75} & \textbf{\color[HTML]{0000FF}  -0.02} & \textbf{\color[HTML]{0000FF} 0.26} & \textbf{\color[HTML]{0000FF} -0.02} & \textbf{\color[HTML]{0000FF} 0.90}  \\
CarButton2 & {\color[HTML]{656565} 0.11} & {\color[HTML]{656565} 5.32} & {\color[HTML]{656565} -0.02} & {\color[HTML]{656565} 1.12} & {\color[HTML]{656565} 0.23} & {\color[HTML]{656565} 12.87} & \textbf{-0.03} & \textbf{0.97} & \textbf{0.01} & \textbf{0.58} & \textbf{\color[HTML]{0000FF} 0.04} & \textbf{\color[HTML]{0000FF} 0.75} \\
CarPush1 & {\color[HTML]{656565} 0.21} & {\color[HTML]{656565} 3.66} & {\color[HTML]{656565} 0.15} & {\color[HTML]{656565} 1.34} & {\color[HTML]{656565} 0.27} & {\color[HTML]{656565} 2.12} & {\color[HTML]{656565} 0.26} & {\color[HTML]{656565} 1.03} & \textbf{\color[HTML]{0000FF} 0.28} & \textbf{\color[HTML]{0000FF} 0.28} & \textbf{0.24} & \textbf{0.50} \\
CarPush2 & {\color[HTML]{656565} 0.11} & {\color[HTML]{656565} 2.96} & \textbf{0.05} & \textbf{0.93} & {\color[HTML]{656565} 0.16} & {\color[HTML]{656565} 4.60} & {\color[HTML]{656565} 0.12} & {\color[HTML]{656565} 2.65} & \textbf{\color[HTML]{0000FF} 0.14} & \textbf{\color[HTML]{0000FF} 0.89} & \textbf{0.05} & \textbf{0.77} \\
CarGoal1 & {\color[HTML]{656565} 0.40} & {\color[HTML]{656565} 4.22} & {\color[HTML]{656565} 0.23} & {\color[HTML]{656565} 1.03} & {\color[HTML]{656565} 0.60} & {\color[HTML]{656565} 3.15} & {\color[HTML]{656565} 0.41} & {\color[HTML]{656565} 1.16} & \textbf{\color[HTML]{0000FF}  0.49} & \textbf{\color[HTML]{0000FF} 0.83} & \textbf{0.28} & \textbf{0.62} \\
CarGoal2 & {\color[HTML]{656565} 0.34} & {\color[HTML]{656565} 3.67} & {\color[HTML]{656565} 0.15} & {\color[HTML]{656565} 2.35} & {\color[HTML]{656565} 0.45} & {\color[HTML]{656565} 6.05} & {\color[HTML]{656565} 0.13} & {\color[HTML]{656565} 1.16} & \textbf{0.06} & \textbf{0.33} & \textbf{\color[HTML]{0000FF} 0.11} & \textbf{\color[HTML]{0000FF} 0.59} \\
AntVel & {\color[HTML]{656565} 0.98} & {\color[HTML]{656565} 33.12} & {\color[HTML]{656565} 0.68} & {\color[HTML]{656565} 2.16} & \textbf{\color[HTML]{0000FF} 0.98} & \textbf{\color[HTML]{0000FF} 0.91} & \textbf{0.31} & \textbf{0.00} & \textbf{0.89} & \textbf{0.00} & \textbf{0.91} & \textbf{0.58} \\
HalfCheetahVel & {\color[HTML]{656565} 0.93} & {\color[HTML]{656565} 18.73} & \textbf{0.73} & \textbf{0.25} & \textbf{\color[HTML]{0000FF} 0.97} & \textbf{\color[HTML]{0000FF} 0.55} & \textbf{0.87} & \textbf{0.23} & \textbf{0.89} & \textbf{0.00} & \textbf{0.87} & \textbf{0.02} \\
SwimmerVel & {\color[HTML]{656565} 0.45} & {\color[HTML]{656565} 15.08} & \textbf{\color[HTML]{0000FF} 0.45} & \textbf{\color[HTML]{0000FF} 0.82} & {\color[HTML]{656565} 0.67} & {\color[HTML]{656565} 1.47} & {\color[HTML]{656565} 0.42} & {\color[HTML]{656565} 0.31} & \textbf{-0.04} & \textbf{0.00} & \textbf{0.12} & \textbf{0.84} \\
\midrule
\textbf{SafetyGym Average} & {\color[HTML]{656565} 0.41} & {\color[HTML]{656565} 11.25} & {\color[HTML]{656565} 0.27} & {\color[HTML]{656565} 1.69} & {\color[HTML]{656565} 0.50} & {\color[HTML]{656565} 4.31} & {\color[HTML]{656565} 0.28} & {\color[HTML]{656565} 1.36} & \textbf{\color[HTML]{0000FF} 0.30} & \textbf{\color[HTML]{0000FF} 0.35} & \textbf{0.29} & \textbf{0.62} \\
\midrule
AntRun & {\color[HTML]{656565} 0.80} & {\color[HTML]{656565} 17.31} & {\color[HTML]{656565} 0.61} & {\color[HTML]{656565} 1.51} & {\color[HTML]{656565} 0.70} & {\color[HTML]{656565} 1.88} & {\color[HTML]{656565} 0.63} & {\color[HTML]{656565} 5.43} & \textbf{0.45} & \textbf{0.03} & \textbf{\color[HTML]{0000FF} 0.63} & \textbf{\color[HTML]{0000FF} 0.84} \\
BallRun & {\color[HTML]{656565} 0.53} & {\color[HTML]{656565} 10.20} & \textbf{0.18} & \textbf{0.89} & \textbf{\color[HTML]{0000FF} 0.32} & \textbf{\color[HTML]{0000FF} 0.45} & {\color[HTML]{656565} 0.29} & {\color[HTML]{656565} 4.24} & \textbf{0.18} & \textbf{0.00} & \textbf{0.24} & \textbf{0.00} \\
CarRun & {\color[HTML]{656565} 0.90} & {\color[HTML]{656565} 3.37} & \textbf{0.86} & \textbf{0.44} & {\color[HTML]{656565} 0.99} & {\color[HTML]{656565} 1.10} & {\color[HTML]{656565} 0.97} & {\color[HTML]{656565} 1.01} & \textbf{0.73} & \textbf{0.14} & \textbf{\color[HTML]{0000FF} 0.93} & \textbf{\color[HTML]{0000FF} 0.57} \\
DroneRun & {\color[HTML]{656565} 0.60} & {\color[HTML]{656565} 12.08} & {\color[HTML]{656565} 0.48} & {\color[HTML]{656565} 2.75} & \textbf{\color[HTML]{0000FF} 0.58} & \textbf{\color[HTML]{0000FF} 0.30} & {\color[HTML]{656565} 0.59} & {\color[HTML]{656565} 1.41} & \textbf{0.30} &\textbf{0.55} & \textbf{0.55} & \textbf{0.32} \\
AntCircle & {\color[HTML]{656565} 0.55} & {\color[HTML]{656565} 16.89} & {\color[HTML]{656565} 0.41} & {\color[HTML]{656565} 6.04} & {\color[HTML]{656565} 0.48} & {\color[HTML]{656565} 7.44} & {\color[HTML]{656565} 0.37} & {\color[HTML]{656565} 2.50} & \textbf{0.20} & \textbf{0.00} & \textbf{\color[HTML]{0000FF} 0.34} & \textbf{\color[HTML]{0000FF} 0.23} \\
BallCircle & {\color[HTML]{656565} 0.73} & {\color[HTML]{656565} 8.76} & \textbf{0.13} & \textbf{0.28} & {\color[HTML]{656565} 0.68} & {\color[HTML]{656565} 2.10} & {\color[HTML]{656565} 0.63} & {\color[HTML]{656565} 1.89} & \textbf{0.34} & \textbf{0.00} & \textbf{\color[HTML]{0000FF} 0.40} & \textbf{\color[HTML]{0000FF} 0.26} \\
CarCircle & {\color[HTML]{656565} 0.33} & {\color[HTML]{656565} 10.19} & {\color[HTML]{656565} 0.23} & {\color[HTML]{656565} 1.07} & {\color[HTML]{656565} 0.71} & {\color[HTML]{656565} 2.19} & \textbf{0.49} & \textbf{0.73} & \textbf{0.40} & \textbf{0.11} & \textbf{0.21} & \textbf{0.68} \\
DroneCircle & {\color[HTML]{656565} 0.71} & {\color[HTML]{656565} 9.46} & \textbf{0.42} & \textbf{0.60} & {\color[HTML]{656565} 0.55} & {\color[HTML]{656565} 1.29} & {\color[HTML]{656565} 0.54} & {\color[HTML]{656565} 2.36} & \textbf{\color[HTML]{0000FF} 0.48} & \textbf{\color[HTML]{0000FF} 0.00} & \textbf{0.43} & \textbf{0.00} \\
\midrule
\textbf{BulletGym Average} & {\color[HTML]{656565} 0.64} & {\color[HTML]{656565} 11.03} & {\color[HTML]{656565} 0.42} & {\color[HTML]{656565} 1.70} & {\color[HTML]{656565} 0.63} & {\color[HTML]{656565} 2.09} & {\color[HTML]{656565} 0.56} & {\color[HTML]{656565} 2.45} & \textbf{0.39} & \textbf{0.10} & \textbf{\color[HTML]{0000FF} 0.47} & \textbf{\color[HTML]{0000FF} 0.36} \\
\midrule
easysparse & {\color[HTML]{656565} 0.67} & {\color[HTML]{656565} 7.64} & \textbf{0.36} & \textbf{0.00} & \textbf{0.05} & \textbf{0.10} & {\color[HTML]{656565} 0.26} & {\color[HTML]{656565} 6.22} & \textbf{0.34} & \textbf{0.00} & \textbf{\color[HTML]{0000FF} 0.58} & \textbf{\color[HTML]{0000FF} 0.44} \\
easymean & {\color[HTML]{656565} 0.63} & {\color[HTML]{656565} 7.64} & \textbf{0.35} & \textbf{0.00} & \textbf{0.27} & \textbf{0.24} & {\color[HTML]{656565} 0.19} & {\color[HTML]{656565} 4.85} & \textbf{0.38} & \textbf{0.25} & \textbf{\color[HTML]{0000FF} 0.48} & \textbf{\color[HTML]{0000FF} 0.09} \\
easydense & {\color[HTML]{656565} 0.54} & {\color[HTML]{656565} 5.84} & \textbf{0.33} & \textbf{0.00} & {\color[HTML]{656565} 0.43} & {\color[HTML]{656565} 2.31} & {\color[HTML]{656565} 0.26} & {\color[HTML]{656565} 5.81} & \textbf{0.36} & \textbf{0.25} & \textbf{\color[HTML]{0000FF} 0.59} & \textbf{\color[HTML]{0000FF} 0.31}  \\
mediumsparse & {\color[HTML]{656565} 0.82} & {\color[HTML]{656565} 5.25} & \textbf{0.39} & \textbf{0.00} & {\color[HTML]{656565} 0.26} & {\color[HTML]{656565} 2.20} & {\color[HTML]{656565} 0.06} & {\color[HTML]{656565} 1.70} & \textbf{0.42} & \textbf{0.22} & \textbf{\color[HTML]{0000FF} 0.45} & \textbf{\color[HTML]{0000FF} 0.53} \\
mediummean & {\color[HTML]{656565} 0.84} & {\color[HTML]{656565} 4.63} & \textbf{\color[HTML]{0000FF} 0.53} & \textbf{\color[HTML]{0000FF} 0.01} & {\color[HTML]{656565} 0.28} & {\color[HTML]{656565} 2.13} & {\color[HTML]{656565} 0.20} & {\color[HTML]{656565} 1.90} & \textbf{0.39} & \textbf{0.08} & \textbf{0.45} & \textbf{0.53} \\
mediumdense & {\color[HTML]{656565} 0.79} & {\color[HTML]{656565} 4.98} & \textbf{0.35} & \textbf{0.01} & \textbf{0.29} & \textbf{0.77} & {\color[HTML]{656565} 0.03} & {\color[HTML]{656565} 1.18} & \textbf{\color[HTML]{0000FF} 0.49} & \textbf{\color[HTML]{0000FF} 0.44} & \textbf{\color[HTML]{0000FF} 0.49} & \textbf{\color[HTML]{0000FF} 0.03} \\
hardsparse & {\color[HTML]{656565} 0.49} & {\color[HTML]{656565} 7.04} & \textbf{\color[HTML]{0000FF} 0.36} & \textbf{\color[HTML]{0000FF} 0.00} & \textbf{0.17} & \textbf{0.47} & \textbf{0.00} & \textbf{0.82} & \textbf{0.30} & \textbf{0.01} & \textbf{0.25} & \textbf{0.18} \\
hardmean & {\color[HTML]{656565} 0.51} & {\color[HTML]{656565} 5.90} & \textbf{0.25} & \textbf{0.00} & {\color[HTML]{656565} 0.28} & {\color[HTML]{656565} 3.32} & {\color[HTML]{656565} 0.16} & {\color[HTML]{656565} 4.91} & \textbf{0.26} & \textbf{0.09} & \textbf{\color[HTML]{0000FF} 0.31} & \textbf{\color[HTML]{0000FF} 0.47} \\
harddense & {\color[HTML]{656565} 0.41} & {\color[HTML]{656565} 4.75} & \textbf{\color[HTML]{0000FF} 0.34} & \textbf{\color[HTML]{0000FF} 0.00} & {\color[HTML]{656565} 0.24} & {\color[HTML]{656565} 1.49} & {\color[HTML]{656565} 0.02} & {\color[HTML]{656565} 1.21} & \textbf{0.30} & \textbf{0.34} & \textbf{0.21} & \textbf{0.00} \\
\midrule
\textbf{Metadrive Average} & {\color[HTML]{656565} 0.63} & {\color[HTML]{656565} 5.96} & \textbf{0.36} & \textbf{0.00} & {\color[HTML]{656565} 0.25} & {\color[HTML]{656565} 1.45} & {\color[HTML]{656565} 0.13} & {\color[HTML]{656565} 3.18} & \textbf{0.36} & \textbf{0.25} & \textbf{\color[HTML]{0000FF} 0.43} & \textbf{\color[HTML]{0000FF} 0.23} \\
\bottomrule
\end{tabular}
}
\end{sc}
\end{small}
\end{center}
\vspace{-15pt}
\end{table*}

\subsection{Experiments on offline constrained RL}
\label{sec:experiment_on_offline_constrained_rl}

Our main offline constrained RL experiment utilizes the DSRL \citep{liu2024offlinesaferl} benchmark, comparing algorithm performance across three environments: Safety-Gymnasium \citep{safetygym,ji2023safety}, Bullet Safety Gym \citep{gronauer2022bullet}, and MetaDrive \citep{li2022metadrive}. In our setup, the goal is to maximize performance by utilizing as much cost as possible without exceeding the cost limit. Consequently, if both methods successfully stay within the limit, the method that achieves a higher return is preferred.

We compare our \textbf{CORSDICE} with: \textbf{BC-All} (imitates entire dataset), \textbf{BC-Safe}, (imitates safe trajectories), \textbf{BCQ-Lag} (Lagrangian BCQ \citep{fujimoto2019off} with PID controller \citep{stooke2020responsive}, \textbf{BEAR-Lag} (Largrangian BEAR~\citep{kumar2019stabilizing} with a PID controller), and \textbf{COptiDICE}~\citep{lee2021coptidice} (details in \cref{sec:appx_dsrl}).

\cref{tab:exp_dsrl} summarizes the results. CORSDICE was the only algorithm to consistently satisfy the cost constraints across all environments. It outperformed baselines in 27 out of 38 tasks and achieved the highest average performance. Other baseline methods struggled to balance return maximization and constraint satisfaction, frequently violating cost constraints or yielding suboptimal returns. This success is attributed to incorporating accurate off-policy cost evaluation into the semi-gradient DICE methods, which have demonstrated state-of-the-art performance in unconstrained settings. Additional ablation studies on the cost sensitivity are in \cref{sec:appx_ablation}.

\subsubsection{With advanced function approximators}
\label{sec:exp_diffusion}

Inspired by recent research leveraing advanced function approximators like diffusion models \citep{ho2020denoising,song2020score} for enhanced offline RL performance~\citep{chen2021decision,hansen2023idql,wang2022diffusion}, we extend CORSDICE by guiding a behavior-cloned diffusion model with our learned policy correction $w(a|s)$, similar to D-DICE~\citep{mao2024diffusion}. We refer to this extension as \textbf{D-CORSDICE} (details in \cref{sec:appx_d_corsdice}).

We compare D-CORSDICE with: \textbf{D-BC-All} (a behavior-cloning diffusion), \textbf{D-BC-Safe} (diffusion BC on safe trajectories), \textbf{Constrained Decision Transformer (CDT)} \citep{liu2023constrained}, \textbf{TREBI} \citep{lin2023safe}, and \textbf{FISOR} \citep{zheng2024safe}. We strictly follow the experimental setup of \citet{lin2023safe}, utilizing a single, tighter cost threshold compared to the standard DSRL benchmark (details in \cref{sec:appx_dsrl}).

\cref{tab:exp_adv_dsrl} summarizes the results. D-CORSDICE and FISOR were the only methods that consistently satisfied cost constraints across all tasks. While their performance was comparable in Safety Gymnasium, D-CORSDICE achieved superior average performance across all evaluated environments.

\section{Conclusion}

We aimed to extend semi-gradient-based DICE methods, known for their strong performance in offline RL, to the constrained setting.
Yet, our findings revealed that semi-gradient DICE algorithms are fundamentally unable to perform policy evaluation, as they produce policy corrections instead of stationary distribution corrections, making their extension to the constrained setting non-trivial. To address this limitation, we proposed a method to recover stationary distribution corrections, and introduced CORSDICE, a novel offline constrained RL algorithm that outperforms existing baselines.

While this work focuses specifically on constrained RL problems, the proposed method can be readily applied to enhance other DICE-based algorithms for problems requiring OPE, such as ROI maximization~\citep{kimroidice}.

\paragraph{Limitations}

Extracting valid stationary distribution requires an additional approximator in our method, potentially leading to increased training complexity. However, we contend that this cost is surpassed by the benefits. Furthermore, due to innate limitations of offline RL, performance may vary depending on the chosen regularization.

\bibliography{neurips_2024}
\bibliographystyle{plainnat}


\appendix

\section{OptiDICE and COptiDICE}
\label{Appendix-A}
In this section, we provide full derivation of OptiDICE \cite{lee2021optidice} and COptiDICE \cite{lee2021coptidice}.
\subsection{OptiDICE}
\label{Appendix-A.1}
We begin with the convex optimization problem \eqref{eq:OptiDICE-whole} OptiDICE solves.
\begin{align*}
    \max_{d\geq 0}\;& \mathbb{E}_{(s,a)\sim d}[r(s,a)] - \alpha \sum_{s,a}d_D(s,a)f\left(\frac{d(s,a)}{d_D(s,a)}\right)\\
    \text{s.t.}\;&(1-\gamma)p_0(s)
    = \sum_{a}d(s,a) - \gamma(\mathcal{T}_{*}d)(s)\;\forall s
\end{align*}
where $(\mathcal{T}_*d)(s):=\sum_{\bar{s},\bar{a}}T(s\mid \bar{s},\bar{a})d(\bar{s},\bar{a})$. For simplicity of derivation, we reformulate the optimization problem in terms of the stationary distribution correction $w(s,a)=d(s,a)/d_D(s,a)$.
\begin{align*}
    \max_{w\geq 0}\;& \mathbb{E}_{(s,a)\sim d_D}[w(s,a)r(s,a)] - \alpha \sum_{s,a}d_D(s,a)f\left(w(s,a)\right)\\
    \text{s.t.}\;&(1-\gamma)p_0(s)
    = \sum_{a}w(s,a)d_D(s,a) - \gamma(\mathcal{T}_{*}d_w)(s)\;\forall s
\end{align*}
where $(\mathcal{T}_*d_w)(s):=\sum_{\bar{s},\bar{a}}T(s\mid \bar{s},\bar{a})w(\bar{s},\bar{a})d_D(\bar{s},\bar{a})$.

We obtain Lagrangian dual $\max_{w \ge 0}\min_{\nu}\mathcal{L}(w,\nu)$ of the reformulated problem where $\nu(s)$ is a Lagrangian multiplier for the Bellman flow constraint.
\begin{align}
\mathcal{L}(w,\nu):=&\mathbb{E}_{(s,a)\sim d_D}\left[w(s,a)r(s,a)-\alpha f\left(w(s,a)\right)\right]\\
&+\sum_s\nu(s)\left((1-\gamma)p_0(s)+\gamma (\mathcal{T}_{*}(d_w)(s))-\sum_{a}w(s,a)d_D(s,a)\right) \nonumber\\=&\mathbb{E}_{(s,a)\sim d_D} \left[w(s,a)\left(r(s,a) + \gamma \sum_{s'}T(s'|s,a)\nu(s') - \nu(s)\right)-\alpha f\left(w(s,a)\right)\right]\\
&+\mathbb{E}_{s_0\sim p_0} [(1-\gamma)\nu(s_0)]\nonumber\\
=&\mathbb{E}_{(s,a)\sim d_D} \left[w(s,a)e_{\nu}(s,a)-\alpha f\left(w(s,a)\right)\right]+\mathbb{E}_{s_0\sim p_0} [(1-\gamma)\nu(s_0)] \label{appendix-optidice-Lwv}
\end{align}
where $\sum_s \nu(s) (\mathcal{T}_{*}d_w)(s) = \sum_{s,a}w(s,a)d_D(s,a)\sum_{s'}T(s'|s,a)\nu(s')$ and $e_{\nu}(s,a) = r(s,a) + \gamma \sum_{s'}T(s'|s,a)\nu(s') - \nu(s)$.

Due to the convexity of the problem, strong duality can be established via Slater's condition. We follow the assumption in \cite{lee2021optidice} that all states are reachable within a given MDP. This assumption ensures the strict feasibility of $d(s,a)>0, \ \forall s,a$, thereby satisfying Slater's condition. The strong duality allows the optimization order to be switched as shown below.
\begin{align}
    \max_{w \ge 0}\min_{\nu}\mathcal{L}(w,\nu)=\min_{\nu}\max_{w \ge 0}\mathcal{L}(w,\nu)
\end{align}
The reordering enables inner maximization over $w(s,a)$, whose optimal solution satisfies $\frac{\partial\mathcal{L}(w,\nu)}{\partial w(s,a)}=0 \ \forall s,a$. Optimal $w^{*}_{\nu}(s,a)$ can be expressed in a closed form in terms of $\nu$.
\begin{align}
     w_{\nu}^{*}(s,a) \ = \max\left(0,(f')^{-1}\left(\frac{e_{\nu}(s,a)}{\alpha}\right)\right) \label{appendix-optidice-w}
 \end{align}
When $w_{\nu}^{*}(s,a)$ is plugged into the dual function \eqref{appendix-optidice-Lwv}, $\nu$ loss of OptiDICE is expressed as,
\begin{align}
    \min_{\nu}\mathcal{L}(w_{\nu}^{*},\nu) &= \mathbb{E}_{s_0\sim p_0} [(1-\gamma)\nu(s_0)]
     +\mathbb{E}_{(s,a)\sim d_D} \left[w_{\nu}^{*}(s,a)e_{\nu}(s,a)-\alpha f\left(w_{\nu}^{*}(s,a)\right)\right]\nonumber\\&=\mathbb{E}_{s_0\sim p_0} [(1-\gamma)\nu(s_0)]
     +\alpha\mathbb{E}_{(s,a)\sim d_D} \left[f^{*}_{0}\left( \frac{e_{\nu}(s,a)}{\alpha}\right)\right]\label{appendix-a-nuloss}
\end{align}
where $f^*_0(y):=\max_{x\geq0}xy-f(x)$ is a convex conjugate of $f$ in $\mathbb{R}^+$.

\paragraph{Policy extration}
After obtaining the solution $\nu^{*}$, we need to extract a policy $\pi^*$ that induces the stationary distribution $w^{*}_{\nu^*}(s,a)d_D(s,a)$. In finite domains, the policy can be computed by $\pi^{*}(a|s)=\frac{w(s,a)d_D(s,a)}{\sum_a w(s,a)d_D(s,a)}$. In continuous domains, assuming a parameterized policy $\pi_{\theta}$, we adopt weighted behavior cloning by minimizing:
\begin{align}
    \min_{\pi_{\theta}} \; - \mathbb{E}_{(s,a)\sim d_D}[w^*_{\nu^*}(s,a)\log \pi_{\theta}(a|s)] 
\end{align}

\subsection{COptiDICE}
\label{Appendix-A.2}
COptiDICE is a constrained version of OptiDICE, where the following constraint is added to the convex optimization \eqref{eq:OptiDICE-whole}:
\begin{align*}
\sum_{s,a}d(s,a)c(s,a) \leq (1-\gamma)C_\textrm{lim}=:\tilde{C}_\textrm{lim}.
\end{align*}

This results in COptiDICE solving offline constrained RL problem defined as:
\begin{align*}
    \max_{w\geq 0}\;& \mathbb{E}_{(s,a)\sim d_D}[w(s,a)r(s,a)] - \alpha \sum_{s,a}d_D(s,a)f\left(w(s,a)\right)\\
    \text{s.t.}\;&(1-\gamma)p_0(s)
    = \sum_{a}w(s,a)d_D(s,a) - \gamma(\mathcal{T}_{*}d_w)(s)\;\forall s\\
    &\sum_{s,a}w(s,a)d_D(s,a)c(s,a)\leq \tilde{C}_\textrm{lim}
\end{align*}

We follow the approach from \cref{Appendix-A.1} to derive the loss funtions of COptiDICE. We obtain Lagrangian dual $\max_{w \ge 0}\min_{\nu, \lambda\geq0}\mathcal{L}(w,\nu,\lambda)$ of the reformulated problem where $\lambda$ is additionally introduced as a Lagrangian multiplier for the cost constraint.
\begin{align}
\mathcal{L}(w,\nu,\lambda):=&\mathbb{E}_{(s,a)\sim d_D}\left[w(s,a)(r(s,a)-\lambda c(s,a)-\alpha f\left(w(s,a)\right)\right] \nonumber \\ &+ \sum_s\nu(s)\left((1-\gamma)p_0(s)+\gamma (\mathcal{T}_{*}(d_w)(s))-\sum_{a}w(s,a)d_D(s,a)\right) +\lambda \tilde{C}_\textrm{lim} \label{appendix-coptidice-Lwv-0}\\
     =&\mathbb{E}_{(s,a)\sim d_D} \left[w(s,a)e_{\nu,\lambda}(s,a)-\alpha f\left(w(s,a)\right)\right]+\mathbb{E}_{s_0\sim p_0} [(1-\gamma)\nu(s_0)]+\lambda \tilde{C}_\textrm{lim} \label{appendix-coptidice-Lwv}
 \end{align}
 where $e_{\nu,\lambda}(s,a) = r(s,a)-\lambda c(s,a) + \gamma \sum_{s'}T(s'|s,a)\nu(s') - \nu(s)$.
 
 The reordering based on strong duality enables inner maximization over $w(s,a)$, whose optimal solution satisfies $\frac{\partial\mathcal{L}(w,\nu,\lambda)}{\partial w(s,a)}=0 \ \forall s,a$. Optimal $w^{*}_{\nu,\lambda}(s,a)$ can be expressed in a closed form in terms of $\nu$ and $\lambda$.
\begin{align}
     w_{\nu,\lambda}^{*}(s,a) \ = \max\left(0,(f')^{-1}\left(\frac{e_{\nu,\lambda}(s,a)}{\alpha}\right)\right) \label{appendix-coptidice-w}
 \end{align}
When $w_{\nu,\lambda}^{*}(s,a)$ is plugged into the dual function \eqref{appendix-coptidice-Lwv}, $\nu$ loss and $\lambda$ loss of COptiDICE can be expressed as,
\begin{align*}
    \min_{\nu}\; &\mathbb{E}_{s_0\sim p_0} [(1-\gamma)\nu(s_0)]
     +\mathbb{E}_{(s,a)\sim d_D} \left[w_{\nu,\lambda}^{*}(s,a)e_{\nu,\lambda}(s,a)-\alpha f\left(w_{\nu,\lambda}^{*}(s,a)\right)\right]\\&=\mathbb{E}_{s_0\sim p_0} [(1-\gamma)\nu(s_0)]
     +\alpha\mathbb{E}_{(s,a)\sim d_D} \left[f^{*}_{0}\left( \frac{e_{\nu,\lambda}(s,a)}{\alpha}\right)\right]
\end{align*}
 We derive $\lambda$ loss from \eqref{appendix-coptidice-Lwv-0} to emphasize its role as a Lagrangian multiplier that ensures the satisfaction of the cost constraint:
\begin{align*}
     \min_{\lambda\geq0}\;\lambda \left(\tilde{C}_\textrm{lim} - \mathbb{E}_{(s,a)\sim d_{D}}[w^{*}_{\nu, \lambda}(s,a)c(s,a)]\right)
\end{align*}

\section{SemiDICE}
\label{appx:semi_dice}
In this section, we derive the semi-gradient variants of OptiDICE (SemiDICE, f-DVL and ODICE) and clarify the characteristics of their optimal solution. We show that SemiDICE returns a valid policy correction rather than a stationary distribution correction, while showing f-DVL and ODICE often violates the validity conditions as a policy correction and a stationary distribution correction as depicted in \cref{fig:random_mdp}. This property makes SemiDICE suitable for CORSDICE framework as it requires a valid policy correction that satisfies $\sum_{a}w(s,a)\pi_D(a|s)=1$. While various semi-gradient losses such as dual-V and $f$-DVL were introduced in \cite{sikchi2023dual}, we derive SemiDICE due to subtle differences in the loss functions and derivations.

As mentioned in our paper, prior semi-gradient methods have involved three modifications: (1) partially or entirely omitting the gradient from next state $\nu(s')$ in $e_\nu(s,a)$, (2) replacing the initial state distribution, $p_0(s)$, with the dataset distribution, and (3) introducing a hyperparameter $\beta$ to balance loss terms while removing the hyperparameter $\alpha$. We divide the sections based on (1) to separately analyze semi-gradient method that (partially/entirely) omits the gradient from next state $\nu(s')$ in $e_\nu(s,a)$ of $\nu$ loss \eqref{appendix-a-nuloss} of OptiDICE.

\subsection{Semi-gradient DICE algorithms that entirely omit the gradient from the next state}
\label{appx:semi_dice_derivation}

We give two semi-gradient DICE algorithms that entirely omit the gradient from the next state $\nu(s')$: SemiDICE and $f$-DVL~\citep{sikchi2023dual}. They share a common characteristic where the term $r(s,a)+\gamma \sum_{s'}T(s'|s,a) \nu(s')$ within $e_{\nu}(s,a)$ is separately estimated by an additional function approximator $Q(s,a)$ with $Q$ loss given below:
\begin{align*}
\min_{Q}\;\mathbb{E}_{(s,a,s')\sim d_D}[(Q(s,a) - (r + \gamma\nu(s')))^2]
\end{align*}
where the use of $Q(s,a)$ is enabled as the gradient from the next state $\nu(s')$ is completely ignored.

\paragraph{SemiDICE} We first derive our algorithm, SemiDICE, from $\nu$ loss of OptiDICE \eqref{appendix-a-nuloss}. We also provide loss function of $f$-DVL for comparison.
\begin{align}
    \mathcal{L}_{\text{OptiDICE}}(\nu) &= \sum_{s}(1-\gamma)p_0(s)\nu(s)\\
    &+\sum_{s,a}d_D(s,a)\left[\alpha f^{*}_{0}\left( \frac{r(s,a) + \gamma \sum_{s'}T(s'|s,a)\nu(s') - \nu(s)}{\alpha}\right)\right] \label{appendix:optidice}\\
    \mathcal{L}_{\text{f-DVL}}(\nu) &= \sum_{s,a}d_D(s,a)\left[(1-\beta)\nu(s)+\beta f^{*}_{0}\left(Q(s,a) - \nu(s)\right)\right] \label{appendix:fdvl}\\
    \mathcal{L}_{\text{SemiDICE}}(\nu) &=\sum_{s,a}d_D(s,a)\left[\nu(s)+\alpha f^{*}_{0}\left( \frac{Q(s,a) - \nu(s)}{\alpha}\right)\right]\label{appendix-semidice}
\end{align}

In \cite{sikchi2023dual}, $f$-DVL applies the semi-gradient technique to OptiDICE and simply replaces the initial state distribution $p_0(s)$ with the dataset distribution $d_D(s)$. However, we propose an alternative interpretation to demonstrate that the replacement can also be understood as a semi-gradient approach. To establish this, we introduce a minor assumption: the dataset policy $\pi_D(a|s)$ and the policy being optimized $\pi(a|s)$ share the same initial state distribution $p_0(s)$.

Under this assumption, we can replace $(1-\gamma)p_0(s)$ with $-\gamma(\mathcal{T}_{*}d_D)(s)+\sum_{a}d_D(s,a)$. This substitution is justified because the stationary distribution of the dataset policy, $d_D(s,a)$, satisfies the Bellman flow constraint as well. This replacement does not break the Bellman flow constraint, as long as the assumption holds. Consequently, we extend this relationship to the equality shown below:
\begin{align}
    \sum_{s}(1-\gamma)p_0(s)\nu(s) &= \sum_{s}\nu(s)\left(-\gamma(\mathcal{T}_{*}d_D)(s)+\sum_{a}d_D(s,a)\right)\\
    &=\sum_{s,a}d_D(s,a)\left(-\gamma \sum_{s'}T(s'|s,a)\nu(s')+ \nu(s)\right)
\end{align}
We apply this relationship to rewrite the Lagrangian dual $\mathcal{L}(w,\nu)$ \eqref{appendix-optidice-Lwv} of OptiDICE.
\begin{align*}
\max_{w\geq 0}\min_{\nu}\mathcal{L}(w,\nu)=&\mathbb{E}_{s\sim p_0}[\nu(s)] + \mathbb{E}_{(s,a)\sim d_D}\left[w(s,a)e_{\nu}(s,a)-\alpha f\left(w(s,a)\right)\right]\\
     =&\mathbb{E}_{(s,a)\sim d_D} [- \gamma \sum_{s'}T(s'|s,a)\nu(s') + \nu(s)]\\
     &+ \mathbb{E}_{(s,a)\sim d_D} \left[w(s,a)e_{\nu}(s,a)-\alpha f\left(w(s,a)\right)\right]
\end{align*}
We follow the same derivation from OptiDICE (\cref{Appendix-A}) to obtain the closed from solution of $w^{*}(s,a)$ that satisfies $\frac{\partial\mathcal{L}(w,\nu)}{\partial w(s,a)}=0$. Since the term with initial distribution is independent to the maximization of $w(s,a)$, its closed form solution is equivalent to that of OptiDICE \eqref{appendix-optidice-w}.
\begin{align*}
    w^{*}_{\nu}(s,a) = \max\left(0,(f')^{-1}\left(\frac{r(s,a)+\gamma \sum_{s'} T(s'|s,a)\nu(s') - \nu(s)}{\alpha}\right)\right)
\end{align*}
We use the closed form solution $w^{*}(s,a)$ to derive $\nu$ loss without the initial state distribution $p_0(s)$.
\begin{align}
    \min_{\nu} \; &\sum_{s,a}d_D(s,a)\left[-\gamma \sum_{s'}T(s'|s,a)\nu(s')\right]\nonumber\\
    &+\sum_{s,a}d_D(s,a)\left[\nu(s)+\alpha f^{*}_{0}\left( \frac{r(s,a) + \gamma \sum_{s'}T(s'|s,a)\nu(s') - \nu(s)}{\alpha}\right)\right]\label{appendix-B-semidice-deriv}
\end{align}
At this point, we apply the semi-gradient technique to  neglect the gradients from $\nu(s')$ and approximate $Q(s,a)$ with $r(s,a)+\sum_{s'}T(s'|s,a)\nu(s')$. We note that the gradients $\nu(s')$ were ignored both inside and outside the convex function $f^{*}_{0}(x)$.
\begin{align*}
\mathcal{L}_{\text{SemiDICE}}(\nu) &= \sum_{s,a}d_D(s,a)\left[ \nu(s)+\alpha f^{*}_{0}\left( \frac{Q(s,a) - \nu(s)}{\alpha}\right)\right]\\
    w^{*}(s,a)&= \max\left(0,(f')^{-1}\left(\frac{Q(s,a) - \nu(s)}{\alpha}\right)\right)
\end{align*}
The semi-gradient technique causes $\nu$ to lose its role as a Lagrangian multiplier that ensures the satisfaction of the Bellman flow constraints of $w^{*}(s,a)d_D(s,a)$. \cref{prop:solution_of_semi_dice} states that the optimal solution of SemiDICE is a policy correction rather than a valid stationary distribution correction and we provide its proof in the following paragraph.

\paragraph{\cref{prop:solution_of_semi_dice}}
 The correction $w^{*}(s,a)$ obtained by the optimal $\nu^*=\arg\min_\nu \mathbb{E}_{d_D}\big[\nu(s)+\alpha f^*_0(\tfrac{Q(s,a) - \nu(s)}{\alpha})\big]$,
\begin{equation*}
w^{*}(s,a)=\max\big(0,(f')^{-1}\big( \tfrac{Q(s,a)-\nu^{*}(s)}{\alpha} \big)\big),
\end{equation*} violates the Bellman flow constraint \eqref{eq:bellman_flow} but satisfies the following conditions for $w^{*}(s,a)$ to act as a policy correction:
\begin{align*}
\sum_a w^{*}(s,a)\pi_D(a|s) = 1, \ w^{*}(s,a) \geq 0, \ \forall s,a.
\end{align*}

\begin{proof}
The derivative of the $\nu$ loss $\mathcal{L}_{\text{SemiDICE}}(\nu)$ w.r.t. $\nu(s)$ is given as
\begin{align*}
    \frac{\partial \mathcal{L}_{\text{SemiDICE}}(\nu)}{\partial \nu(s)} &= \sum_a d_D(s,a)\left(1-(f^{*}_{0})'\left(\frac{Q(s,a)-\nu(s)}{\alpha}\right) \right) \nonumber\\
    &= \sum_a d_D(s,a)\left(1-\max\left(0,(f')^{-1}\left(\frac{Q(s,a)-\nu(s)}{\alpha}\right) \right)\right) \nonumber
\end{align*}
where $(f^{*}_{0})'(x) = \max(0,(f')^{-1}(x))$. 

Due to the convexity of $f^{*}_{0}(x)$, optimal $w^{*}(s,a)=\max\left(0,(f')^{-1}\left(\frac{Q(s,a)-\nu^{*}(s)}{\alpha}\right) \right)$ obtained from $\nu^{*}$ that satisfies $\partial \mathcal{L}_{\text{SemiDICE}}(\nu)/\partial \nu = 0$ satisfies the equality below.
\begin{align*}
    &\sum_a d_D(s,a)\left(1-w^{*}(s,a)\right)=0\\
    &\sum_a w^{*}(s,a)d_D(s,a) = \sum_{a}d_D(s,a)\\
    &\sum_a w^{*}(s,a)\pi_D(a|s) = 1  \ \ \forall s
\end{align*}
where state stationary distribution $d_D(s)$ is divided in each sides and $\sum_a\pi_D(a|s)=1$. This indicates that weighted stationary distribution $w^{*}(s,a)d_D(s,a)$ is not a valid stationary distribution, while $w^{*}(s,a)\pi_D(a|s)$ is a policy, as its marginal sum over actions is 1.
\end{proof}

\paragraph{$f$-DVL} Based on the derivation of SemiDICE, we identify the optimal solution of $f$-DVL. $\nu$ loss of $f$-DVL $\mathcal{L}_{\text{f-DVL}}$, is simply derived from $\nu$ loss of SemiDICE by reweighting the terms with $\beta$ instead of $\alpha$. $\mathcal{L}_{\text{f-DVL}}$ and its optimal weight $w^{*}(s,a)$ is given as,
\begin{align*}
    \mathcal{L}_{\text{f-DVL}}(\nu) &= \sum_{s,a}d_D(s,a)\left[ (1-\beta)\nu(s)+\beta f^{*}_{0}\left( Q(s,a) - \nu(s)\right)\right]\\
    w^{*}_{\text{f-DVL}}(s,a)&= \max\left(0,(f')^{-1}\left(Q(s,a) - \nu(s)\right)\right)
\end{align*}

We also show the property of the optimal $w^{*}_{\text{f-DVL}}(s,a)$ from the derivative of $\mathcal{L}_{\text{f-DVL}}$ w.r.t. $\nu(s)$ given as,
\begin{align*}
    \frac{\partial \mathcal{L}_{\text{f-DVL}}}{\partial \nu(s)} &= \sum_a d_D(s,a)\left(1-\beta + \beta (f^{*}_{0})'\left(Q(s,a)-\nu(s)\right) \right) \nonumber\\
    &= \sum_a d_D(s,a)\left(1-\beta + \beta \max\left(0,(f')^{-1}\left(Q(s,a)-\nu(s)\right)\right)\right) \nonumber\\
    &= \sum_a d_D(s,a)\left(1-\beta + \beta w^{*}_{\text{f-DVL}}(s,a)\right)\nonumber
\end{align*}
Optimal correction $w^{*}_{\text{f-DVL}}$ is obtained from the optimal $\nu^{*}$ that minimizes $\mathcal{L}_{\text{f-DVL}}$.
\begin{align*}
    \beta\sum_a &w_{\text{f-DVL}}^{*}(s,a)d_D(s,a)=(1-\beta)\sum_{a}d_D(s,a) \ \forall s\\
    \sum_a &w_{\text{f-DVL}}^{*}(s,a)\pi_D(a|s) = \frac{1-\beta}{\beta}  \ \ \forall s
\end{align*}
This indicates that the weighted stationary distribution $w^{}_{\text{f-DVL}}(s,a)d_D(a,s)$ is not a valid stationary distribution, whereas $w^{*}_{\text{f-DVL}}(s,a)\pi_D(a|s)$ represents a scaled policy. The weight from $f$-DVL, $w^{*}_{\text{f-DVL}}(s,a)$, is a valid policy correction only when $\lambda=0.5$. While the invalidity of $f$-DVL as a policy correction does not impact its offline RL performance, it affects our off-policy evaluation method within \cref{Section5-extraction} as it requires a valid policy correction. Therefore, we adopt SemiDICE in our formulation of CORSDICE.

\subsection{Semi-gradient DICE algorithms that partially omits the gradient from the next state}
\label{Appendix-B.2}
We give a semi-gradient DICE algorithm that partially omits the gradient from the next state $\nu(s')$: ODICE~\citep{Mao2024odice}. ODICE is an offline RL algorithm built upon $f$-DVL where the gradient from the next state $\nu(s')$ is projected to be orthogonal to the gradient from the current state $\nu(s)$ rather than completely ignoring it.

\paragraph{ODICE} We demonstrate how ODICE is derived by applying the orthogonal gradient approach to the following loss function similar to $\nu$ loss of $f$-DVL: 
\begin{align}
    \min_{\nu} \; &\sum_{s,a,s'}d_D(s,a,s')\left[(1-\beta)\nu(s)+\beta f^{*}_{0}\left(e_{\nu}(s,a,s')\right)\right]\label{odice-1}
\end{align}
where $e_{\nu}(s,a,s')=r(s,a)+\gamma \nu(s')-\nu(s)$. Assuming $\nu$ is parameterized by $\theta$, ODICE addresses the conflict between two gradients within $f^*_0(e_{\nu_{\theta}}(s,a,s'))$: forward gradient $g_{f}$ and backward gradient $g_b$.
\begin{align*}
g_f &= -(f^{*}_{0})'(e_{\nu_{\theta}}(s,a,s'))\nabla_{\theta}\nu_{\theta}(s)\\
g_b &=\gamma(f^{*}_{0})'(e_{\nu_{\theta}}(s,a,s'))\nabla_{\theta}\nu_{\theta}(s')
\end{align*}

The paper claims that the backward gradient may cancel out the effect of the forward gradient, leading to a catastrophic unlearning phenomenon. To address the conflicting gradient issue, the orthogonal gradients are applied to the second term of \eqref{odice-1}.
\begin{align}
\nabla_{\theta} f^{*}_{0}(e_{\nu_{\theta}}(s,a,s')) &= g_f + g_b \\
\nabla_{\text{ortho}}f^{*}_{0}(e_{\nu_{\theta}}(s,a,s')) &=g_f + \eta\left(g_b - \frac{g_b^{T}g_f}{||g_f||^2} g_f\right)
\end{align}
where backward gradient is projected to be orthogonal to the forward gradient, and $\eta$ is a hyperparameter that decides how much the projected gradient is applied.~\citep{Mao2024odice} shows that by setting $\eta$ large enough, orthogonal gradient descent can converge to the same point when full gradient is applied, the first term is an expectation over dataset distribution not an expectation over initial state distribution.

Despite the strong RL performance of ODICE in offline RL, it struggles to converge to a valid stationary distribution correction or policy correction. We identify three key reasons for this issue:
\begin{enumerate}
    \item Difficulty in choosing an appropriate $\eta$: It is challenging to determine a sufficiently large $\eta$ that ensures the equal convergence point of orthogonal and full gradient descent.
    \item Bias in the objective function: ODICE is based on the biased objective \eqref{odice-1}, where the expectation over the transition probability $T$ appears outside the convex function $f^{*}_{0}(x)$. This implies that unless the transition probability is deterministic for all states and actions, the biased objective cannot be directly related to the DICE objectives, where the expectation over $T$ inside $f^{*}(x)$.
    \item Mismatch in the gradients: In \eqref{odice-1}, the first term replaces the initial state distribution $p_0(s)$ with the dataset distribution $d_D$. This substitution can be interpreted as the application of a semi-gradient method, as demonstrated in the derivation of SemiDICE from \eqref{appendix-B-semidice-deriv}. Consequently, even if the second term of \eqref{odice-1} is updated using its full gradient, neither convergence to policy correction nor satisfaction of the Bellman flow constraints is guaranteed.
\end{enumerate}
\cref{fig:random_mdp} provides empirical evidence that ODICE generally fails to converge to either policy correction or stationary distribution correction.

\section{Relationship with Behavior Regularized MDP}
\label{appendix-c}
In this section, we illustrate the close relationship between SemiDICE, SQL \cite{xuoffline}, and XQL \cite{garg2023extreme}, all of which address the same Behavior Regularized MDP using different approximation methods. We first present the optimal solution to the Behavior Regularized MDP introduced in \cite{xuoffline} and elaborate on how practical algorithms are derived from this solution. While SQL and XQL are restricted to specific $f$-divergences (Neyman $\chi^2$-divergence and reverse KL divergence), SemiDICE is not limited to a specific $f$-divergence.

This can be considered an extension of the claim from \cite{sikchi2023dual}—that ``XQL is an instance of semi-gradient DICE with initial state distribution replacement, where the $f$-divergence is the reverse KL divergence"—though we extend this in a different manner. Based on \cref{Section4-analysis} and \cref{appendix-c}, we show SemiDICE is an approximation of the behavior-regularized RL. 

\subsection{Behavior Regularized MDP}

We begin by considering the behavior regularized MDP introduced in \cite{xuoffline}:
\begin{align}
\label{appendix-c-sqlmdp}
\max_{\pi} \ &\mathbb{E}\left[\sum_{t=0}^{\infty}\gamma^{t}\left(r(s_t,a_t)-\alpha f\left( \frac{\pi_D(a_t|s_t)}{\pi(a_t|s_t)}\right)\right)\right]
\end{align}
where the reward is penalized with the $f$-divergence between $\pi_D(a|s)$ and $\pi(a|s)$. In prior works, Neyman $\chi^2$-divergence (SQL) and reverse KL divergence (XQL) between $\pi_D(a|s)$ and $\pi(a|s)$ were employed. We obtain an equivalent MDP by replacing $f(x)$ with $g(x)=xf(1/x)$, as follows:
\begin{align}
\label{appendix-c-ourmdp}
\max_{\pi} \ &\mathbb{E}\left[\sum_{t=0}^{\infty}\gamma^{t}\left(r(s_t,a_t)-\alpha \frac{\pi_D(a_t|s_t)}{\pi(a_t|s_t)} f\left( \frac{\pi(a_t|s_t)}{\pi_D(a_t|s_t)}\right)\right)\right]
\end{align}
This results in reversing the order of the $f$-divergence, where the reward is penalized with the corresponding $f$-divergence between $\pi(a|s)$ and $\pi_D(a|s)$.

\paragraph{Reverse relationship between $f$-divergences} If $D_f(\pi||\pi_D)$ is the $f$-divergence between $\pi$ and $\pi_D$, then $D_g(\pi||\pi_D)$ characterized by $g(x)=xf(1/x)$ is also an $f$-divergence and $D_f(\pi||\pi_D)=D_g(\pi_D||\pi)$. We first show $g(x)$ is a valid function for $f$-divergence if $f(x)$ is valid.
We list three properties that $f(x)$ satisfies as a valid function for $f$-divergence.
\begin{enumerate}
    \item Convexity of $f(x)$ in its domain:
    \begin{itemize}
        \item $f(\theta x + (1-\theta) y)\leq \theta f(x)+ (1-\theta)f(y), \ \text{with} \ 0\leq \theta \leq 1, \ \forall x,y \in \text{dom} \ f$ 
    \end{itemize}
    \item $f(1)=0$ and strict convexity of $f(x)$ at 1
    \begin{itemize}
        \item If $\theta x + (1-\theta) y=1$, $f(1)=f(\theta x + (1-\theta) y) < \theta f(x)+ (1-\theta)f(y), \ \text{with} \ 0\leq \theta \leq 1, \ \forall x,y \in \text{dom} \ f$
    \end{itemize}
\end{enumerate}
We show that $g(x)=xf(1/x)$ also satisfies these properties by using the properties of $f$. We define $c= \theta x + (1-\theta) y$.
\begin{enumerate}
    \item Convexity of $g(x)$ in its domain:
    \begin{itemize}
        \item $g(\theta x + (1-\theta) y)=cf(1/c)=cf(\frac{\theta x}{c}\frac{1}{x}+\frac{(1-\theta) y}{c}\frac{1}{y})\leq \theta g(x)+ (1-\theta)g(y), \ \text{with} \ 0\leq \theta \leq 1, \ \forall x,y \in \text{dom} \ g$ 
    \end{itemize}
    \item $g(1)=0$ and strict convexity of $g(x)$ at 1
    \begin{itemize}
        \item $g(1)=f(1)=0$
        \item If $c=1$, $g(1)=f(1)=f(\frac{\theta x}{1}\frac{1}{x}+\frac{(1-\theta) y}{1}\frac{1}{y}) < \theta g(x)+ (1-\theta)g(y), \ \text{with} \ 0\leq \theta \leq 1, \ \forall x,y \in \text{dom} \ g$
    \end{itemize}
\end{enumerate}
As $f$-divergence characterized by $g(x)$ is a valid $f$-divergence, we now show their reverse relationship.
\begin{align*}
D_f(\pi||\pi_D) &:= \sum_{s,a}\pi_D(a|s)f\left(\frac{\pi(a|s)}{\pi_D(a|s)}\right)=\sum_{s,a}\pi_D(a|s)\frac{\pi(a|s)}{\pi_D(a|s)}f\left(\frac{\pi_D(a|s)}{\pi(a|s)}\right)=D_g(\pi_D||\pi)
\end{align*}

\subsection{Optimal solution of Behavior Regularized MDP}
\label{appendix-c.2-behavopt}
We provide a proof on \cref{proposition4.2} by deriving the optimal policy $\pi^{*}(a|s)$, and its corresponding value functions for our behavior regularized MDP \eqref{appendix-c-ourmdp}. Following the derivations of \cite{xuoffline}, the policy evaluation operator $\mathcal{T}^{\pi}_{f}$ of the behavior regularized MDP is given by,
\begin{align*}
&(\mathcal{T}^{\pi}_{f}Q)(s,a) := r(s,a) + \gamma \mathbb{E}_{s' \sim T(\cdot|s,a)}[V(s')]\\
&V(s) =\mathbb{E}_{a\sim\pi(\cdot|s)}\left[Q(s,a)-\alpha \frac{\pi_D(a|s)}{\pi(a|s)}f\left(\frac{\pi(a|s)}{\pi_D(a|s)}\right)\right]
\end{align*}

The policy learning objective can be expressed as $\max_{\pi}\mathbb{E}_{s\sim d_D}[V(s)]$, where $D$ denotes the dataset distribution. Accordingly, we formulate a convex optimization problem that optimizes policy $\pi$ by solving $\max_{\pi}\mathbb{E}_{s\sim D}[V(s)]$, given $Q(s,a)$ within the dataset $d_D$. We add the constraints to ensure the optimized policy $\pi(a|s)$ is a valid policy.
\begin{align*}
\max_{\pi}\;&\sum_{s,a}d_D(s)\pi(a|s)Q(s,a)-\alpha \sum_{s,a}d_D(s)\pi_D(a|s)f\left(\frac{\pi(a|s)}{\pi_D(a|s)}\right)\\
\textrm{s.t.}\;&\sum_{a}\pi(a|s)=1 \ \forall s,a\\
&\pi(a|s)\geq0 \ \ \forall s,a
\end{align*}

To solve the convex optimization problem, we derive the Lagrangian dual with Lagrangian multiplier $U(s)$ and $\beta(s,a)$ for each policy constraints.
\begin{align*}
    \max_{\pi}\min_{U,\beta \geq 0} \ L(\pi,U,\beta)=&\sum_{s,a}d_D(s)\pi(a|s)Q(s,a)-\alpha\sum_{s}d_D(s)\pi_D(a|s)f\left(\frac{\pi(a|s)}{\pi_D(a|s)}\right)\\
    &-\sum_{s}d_D(s)\sum_{a}U(s)\left(\pi(a|s)-1\right)-\beta(s,a)\pi(a|s)
\end{align*}
The KKT conditions of the problem are as follows,
\begin{align}
    &\pi^{*}(a|s)\geq0,\ \forall s,a  \ \text{and} \ \sum_{a}\pi^{*}(a|s)=1, \ \forall s \nonumber \\
    &\beta^{*}(s,a)\geq 0 , \ \forall s,a \nonumber \\
    &\beta^{*}(s,a)\pi^{*}(a|s)=0, \ \forall s,a \nonumber\\
    &Q(s,a)-\alpha f'\left(\frac{\pi^{*}(a|s)}{\pi_D(a|s)}\right) - U^{*}(s)-\beta^{*}(s,a)=0, \ \forall s,a \label{appendix-c-stationarity}
\end{align}
Due to the sationarity condition \eqref{appendix-c-stationarity}, the optimal policy correction $w^{*}(a|s)$ given $U^{*}(s)$ is
\begin{align}
    \pi^{*}(a|s) &= \max\left(0,(f')^{-1}\left(\frac{Q(s,a)-U^{*}(s)}{\alpha}\right)\right)\pi_D(a|s), \ \forall s,a \label{appendix-c-policy-ratio}
\end{align}

We emphasize that the optimality of $U^{*}(s)$ is independent to the optimality of $Q(s,a)$. Even if $Q(s,a)$ is not equivalent to the optimal $Q^{*}(s,a)$ of \eqref{appendix-c-ourmdp}, $\pi^{*}(a|s)$ derived from $U^{*}(s)$ is still a valid policy correction, due to the role of $U(s)$ as Lagrangian multiplier.

To formulate a loss function solely on $U(s)$, we switch the order of optimization based on strong duality ($\max_{\pi}\min_{U,\beta \geq 0} \ L(\pi,U,\beta) = \min_{U,\beta \geq 0}\max_{\pi} \ L(\pi,U,\beta)$). Slater's condition for the strong duality is easily satisfied as there exists a policy that satisfies $\pi(a|s)>0 , \forall s,a$. We then insert the optimal solutions $\pi^{*}(a|s)$ and $\beta^{*}(s,a)$ into $L(\pi,U,\beta)$ which results in:
\begin{align}
    \min_{U} \ L(\pi^{*},U,\beta^{*})&= \sum_{s}d_D(s)\left[ U(s) +\sum_{a}\pi^{*}(a|s)(Q(s,a)-U(s))\right]\nonumber\\
    &-\sum_{s}d_D(s)\left[\alpha\sum_{a}\pi_D(a|s)f\left(\frac{\pi^{*}(a|s)}{\pi_D(a|s)}\right)\right]\nonumber\\
    &= \mathbb{E}_{(s,a)\sim d_D}\left[U(s)+\alpha f_0^{*}\left(\frac{Q(s,a)-U(s)}{\alpha}\right)\right]\label{appendix-c-uloss}
\end{align}

The optimal Lagrangian dual $L(\pi^{*},U^{*},\beta^{*})$ is equivalent to the optimal solution $V^{*}(s)=\max_{\pi}\mathbb{E}_{s\sim D}[V(s)]$ given $Q(s,a)$.
\begin{align*}
V^{*}(s) = U^{*}(s)+\mathbb{E}_{a\sim\pi_D(\cdot|s)}\left[\alpha f_0^{*} \Big(\frac{Q(s,a)-U^{*}(s)}{\alpha} \Big) \right], \ \forall s
\end{align*}

\paragraph{\cref{proposition4.2}}Therefore, in the behavior regularized MDP \eqref{appendix-c-ourmdp}, the optimal value functions $Q^{*}(s,a)$ and $V^{*}(s)$ and its corresponding optimal policy $\pi^{*}$ satisfy the following optimality conditions for all states and actions.
\begin{subequations}
\label{appendix-c-optimalbehav}
\begin{align}
    U^{*}(s) &= \arg\min_{U(s)} U(s)+\mathbb{E}_{a\sim\pi_D(\cdot|s)}\left[\alpha f_0^{*}(\frac{Q^{*}(s,a)-U(s)}{\alpha}) \right]\label{appendix-c-optimalbehav-b}\\
    V^{*}(s)&= U^{*}(s)+\mathbb{E}_{a\sim\pi_D(\cdot|s)}\left[\alpha f_0^{*}(\frac{Q^{*}(s,a)-U^{*}(s)}{\alpha}) \right]\label{appendix-c-optimalbehav-c}\\
    Q^{*}(s,a)&= r(s,a) + \gamma \mathbb{E}_{s'\sim T(\cdot|s,a)}[V^{*}(s')] \\
    \pi^{*}(a|s) &= \max\left(0,(f')^{-1}\left(\frac{Q^{*}(s,a)-U^{*}(s)}{\alpha}\right)\right)\pi_D(a|s)
\end{align}
\end{subequations}
We now demonstrate how the optimal solution of the behavior regularized MDP \eqref{appendix-c-optimalbehav} is approximated by SemiDICE and SQL. We also demonstrate a special case, XQL, that does not require any approximation.

\paragraph{Approximation in SemiDICE}
We show that SemiDICE approximates the optimal solution of the behavior-regularized MDP \eqref{appendix-c-optimalbehav} by eliminating $V$ and approximating $V^{*}$ with $U^{*}$, i.e., $V^{*}(s) \approx U^{*}(s)$. To elaborate, we give the loss functions and the optimal policy of SemiDICE.
\begin{subequations}
\label{appendix-c-semidice}
\begin{align}
    &\min_{\nu} \; \mathbb{E}_{(s,a)\sim d_D}\left[\nu(s)+\alpha f_0^{*}\left(\frac{Q^{*}(s,a)-\nu(s)}{\alpha}\right) \right]\label{appendix-c-semidice-b}\\
     &\min_{Q} \; \mathbb{E}_{(s,a)\sim d_D}\left[\left(r(s,a)+\gamma\nu(s')-Q(s,a)\right)^2\right]\\
    &\pi^{*}(a|s) = \max\left(0,(f')^{-1}\left(\frac{Q^{*}(s,a)-\nu^{*}(s)}{\alpha}\right)\right)\pi_D(a|s)\label{appendix-c-semidice-a}
\end{align}
\end{subequations}
where $U^{*}$ of the behavior regularized MDP \eqref{appendix-c-optimalbehav-b} and $\nu^{*}(s)$ of \eqref{appendix-c-semidice-b} are equivalent as they converge to same value given $Q(s,a)$. SemiDICE omits the computation of $V$ \eqref{appendix-c-optimalbehav-c} and uses only $\nu$ to update $Q$, which is equivalent to approximating $\mathbb{E}_{a\sim\pi_D(\cdot|s)}\left[\alpha f_0^{*}(\frac{Q^{*}(s,a)-U^{*}(s)}{\alpha}) \right]$ with 0. This indicates that optimal $Q^{*}(s,a)$ of SemiDICE is an approximation of optimal $Q^{*}(s,a)$ of behavior regularized MDP. However, we emphasize that \eqref{appendix-c-semidice-a} is still a valid policy correction as optimization on $\nu$ acted equivalently to the Lagrangian multiplier $U$ that ensures the satisfaction of policy constraints.

\paragraph{Approximation in SQL} We show that SQL approximates the optimal solution of the behavior-regularized MDP \eqref{appendix-c-optimalbehav} by eliminating $U$ and approximating $U^{*}$ with $V^{*}$, i.e., $U^{*}(s) \approx V^{*}(s)-\alpha$. Before describing the approximation, we first apply the $f$-divergence used in SQL to the behavior regularized MDP \eqref{appendix-c-optimalbehav}: Neyman $\chi^2$-divergence between $\pi_D(a|s)$ and $\pi(a|s)$ ($g(x)=1/x+1$), which is equivalent to $\chi^2$-divergence between $\pi(a|s)$ and $\pi_D(a|s)$ ($f(x)=x^2-x$).
\begin{subequations}
\label{appendix-c-SQL}
\begin{align}
    \pi^{*}(a|s) &= \max\left(0,\frac{1}{2}+\frac{Q^{*}(s,a)-U^{*}(s)}{2\alpha}\right)\pi_D(a|s)\label{appendix-c-SQL-a}\\
    U^{*}(s) &= \arg\min_{U(s)} U(s)\nonumber\\
    &+\mathbb{E}_{a\sim\pi_D(\cdot|s)}\left[\alpha \max\left(0, \frac{1}{2}+\frac{Q^{*}(s,a)-U(s)}{2\alpha} \right)\left(\frac{1}{2}+\frac{Q^{*}(s,a)-U(s)}{2\alpha}\right)\right]\label{appendix-c-SQL-b}\\
    V^{*}(s)&= U^{*}(s)\nonumber\\
    &+\mathbb{E}_{a\sim\pi_D(\cdot|s)}\left[\alpha \max\left(0, \frac{1}{2}+\frac{Q^{*}(s,a)-U^{*}(s)}{2\alpha}\right)\left(\frac{1}{2}+\frac{Q^{*}(s,a)-U^{*}(s)}{2\alpha}\right)\right]\label{appendix-c-SQL-c}\\
    Q^{*}(s,a)&= r(s,a) + \gamma \mathbb{E}_{s'\sim T(\cdot|s,a)}[V^{*}(s')]
\end{align}
\end{subequations}
where $f^{*}_{0}(y)=\max\left(0,\frac{1+y}{2}\right)\left(\frac{1+y}{2}\right)$. By applying \eqref{appendix-c-SQL-a} to \eqref{appendix-c-SQL-c}, the following equality is satisfied:
\begin{align*}
    V^{*}(s)&=U^{*}(s)+\alpha\mathbb{E}_{a\sim \pi_D(\cdot|s)}\left[\left(\frac{\pi^{*}(a|s)}{\pi_D(a|s)}\right)^2\right]\approx U^{*}(s)+\alpha
\end{align*}
where the second term is approximated to $\alpha$ in SQL. The approximation leads to the replacement of $U(s)$ within \eqref{appendix-c-SQL-b} with $V(s)-\alpha$, which leads to the loss functions and the optimal policy of SQL given by:
\begin{align*}
    &\min_{V} \; \mathbb{E}_{(s,a) \sim d_D}\left[V(s)+ \alpha \max\left(0,1+\frac{Q^{*}(s,a)-V(s)}{2\alpha}\right) \left(1+\frac{Q^{*}(s,a)-V(s)}{2\alpha}\right)\right]\\
    &\min_{Q} \; \mathbb{E}_{(s,a)\sim d_D}\left[\left(r(s,a)+\gamma\nu(s')-Q(s,a)\right)^2\right]\\
    &\pi^{*}(a|s) = \max\left(0,1+\frac{Q^{*}(s,a)-V^{*}(s)}{2\alpha}\right)\pi_D(a|s)
\end{align*}
However, the approximation causes $V(s)$ of SQL to converge to $U^{*}(s)+\alpha$ from $U^{*}(s)$ \eqref{appendix-c-SQL-b}, rather than $V^{*}(s)$ \eqref{appendix-c-SQL-c}. We introduce two special cases where the equality between $U^{*}(s)+\alpha=V^{*}(s)$ is satisfied. 

\paragraph{Special case in XQL}
 We show that XQL converges to the optimal solution of the behavior-regularized MDP with reverse KL divergence betweeen $\pi_D$ and $\pi$ ($g(x)=-\log x$), which is equivalent to KL-divergence between $\pi$ and $\pi_D$ ($f(x)=x\log x $). We apply the $f$-divergence to the behavior regularized MDP \eqref{appendix-c-optimalbehav}:
\begin{subequations}
\label{appendix-c-XQL}
\begin{align}
    \pi^{*}(a|s) &= \exp\left(\frac{Q^{*}(s,a)-U^{*}(s)}{\alpha}-1\right)\pi_D(a|s)\label{appendix-c-XQL-a}\\
    U^{*}(s) &= \arg\min_{U(s)} U(s)+\mathbb{E}_{a\sim\pi_D(\cdot|s)}\left[\alpha \exp\left(\frac{Q^{*}(s,a)-U(s)}{\alpha}-1\right)\right]\label{appendix-c-XQL-b}\\
    V^{*}(s)&= U^{*}(s)+\mathbb{E}_{a\sim\pi_D(\cdot|s)}\left[\alpha \exp\left(\frac{Q^{*}(s,a)-U^{*}(s)}{\alpha}-1\right)\right]\label{appendix-c-XQL-c}\\
    Q^{*}(s,a)&= r(s,a) + \gamma \mathbb{E}_{s'\sim T(\cdot|s,a)}[V^{*}(s')]
\end{align}
\end{subequations}
where $f^{*}_{0}(y)=\exp\left(y-1\right)$. By applying \eqref{appendix-c-XQL-a} to \eqref{appendix-c-XQL-c}, the following equality is satisfied:
\begin{align*}
    V^{*}(s)&=U^{*}(s)+\alpha\mathbb{E}_{a\sim \pi_D(\cdot|s)}\left[\frac{\pi^{*}(a|s)}{\pi_D(a|s)}\right] = U^{*}(s)+\alpha
\end{align*}
where the approximations of the previous algorithms are not required. The loss functions and the optimal policy of XQL are obtained by substituting $U(s)$ with $V(s)-\alpha$:
\begin{align*}
    &\min_{V} \; \mathbb{E}_{(s,a)\sim d_D}\left[V(s)+\alpha \exp\left(\frac{Q(s,a)-V(s)}{\alpha}\right)\right]\\
    &\min_{Q} \; \mathbb{E}_{(s,a)\sim d_D}\left[\left(r(s,a)+\gamma\nu(s')-Q(s,a)\right)^2\right]\\
    &\pi^{*}(a|s) = \exp\left(\frac{Q^{*}(s,a)-V^{*}(s)}{\alpha}\right)\pi_D(a|s)
\end{align*}
While XQL solves the behvaior regularized MDP with no approximation, the exponential term within $V$ loss makes the algorithm prone to divergence. While the instability can be avoided by adopting high $\alpha$, the regularization becomes to strong and causes its performance of $\pi$ to be bound to $\pi_D$.

\subsection{Proof on SemiDICE avoiding the sparsity problem}
\label{appendix-c.3}
We show SemiDICE and other behavior-regularized does not suffer from the sparsity problem OptiDICE suffers by providing the proof below:
\paragraph{\cref{corollary2}}
Let $w^*$ be the correction optimized by running SemiDICE. There is no state $s$ where $w^{*}(s,a) = 0 \ \forall a$.
\begin{proof}
    Assume there exists a state $s$ whose $w^{*}(s,a)=0 \ \forall a$. The assumption contradicts $\sum_{a}w^{*}(s,a)\pi_D(a|s)=1 \ \forall a$ as $\sum_a w^{*}(s,a)\pi_D(a|s) = 0$ in the state $s$, therefore SemiDICE does not suffer from the sparsity problem.
\end{proof}

\section{State stationary distribution extraction}
\label{Appendix-D}
In this section, we provide a detailed derivation on state stationary extraction (\textbf{Extraction}), where we obtain state stationary distribution correction $w(s)$ induced by policy correction $w(a|s)$. After obtaining $w(s)$, stationary distribution $d(s,a)=w(s)w(a|s)d_D(s,a)$ can be utilized for off-policy cost evaluation in offline constrained RL.
\begin{align*}
\E_{(s,a) \sim d} [ c(s,a) ] = \E_{(s,a)\sim d_{D}}[w(s)w(a|s)c(s,a)]
\end{align*}
We formulate a novel convex optimization problem whose optimal solution corresponds to the state stationary distribution ratio, $w(s)$. We assume that the policy correction, $w(a|s)$, is given:
\begin{subequations}
\label{Appendix-D-optimization}
\begin{align}
    &\max_{w(s)\geq 0}\;-\sum_{s}d_D(s)f\left(w(s)\right)\label{Appendix-D-fdivergence}\\
    &\text{s.t.}\; w(s)d_D(s) = (1-\gamma)p_0(s) + \gamma(\mathcal{T}_{*}d_w)(s)\;\forall s \label{Appendix-D-bellman}
\end{align}
\end{subequations}
where  $(\mathcal{T}_*d_w)(s):=\sum_{\bar{s},\bar{a}}T(s\mid \bar{s},\bar{a})w(\bar{s})w(\bar{a}|\bar{s})d_D(\bar{s},\bar{a})$.

Regardless of the objective, the $|S|$ Bellman flow constraints in the problem is sufficient to uniquely determine $w(s)$. However, the $f$-divergence between $w(s)d_D(s)$ and $d_D(s)$ introduces convexity into optimization, enabling the application of convex optimization and efficient sample-based optimization.

The Lagrangian dual of the problem, with Lagrange multipliers $\mu(s)$ for the constraint \eqref{Appendix-D-bellman}, is given as:
\begin{align}
\max_{w(s)\geq0}\min_{\mu}\;&\mathcal{L}(w,\mu)\\
&:= -\sum_{s}d_D(s)f\left(w(s)\right)+\sum_{s}\mu(s)\left((1-\gamma)p_0(s) + \gamma(\mathcal{T}_{*}d_w)(s)-w(s)d_D(s) \right)\\
&=\sum_{s}(1-\gamma)p_0(s)\mu(s)\nonumber\\
&+ \sum_{s,a}d_D(s,a)\left(w(s)w(a|s)\left(\gamma \sum_{s'}T(s'|s,a)\mu(s') - \mu(s)\right)- f(w(s))\right)\label{Appendix-D-Lag0}
\\
&=(1-\gamma)\mathbb{E}_{s_0 \sim p_0}\left[\mu(s_0)\right]
    + \mathbb{E}_{(s,a)\sim d_D} \left[w(s)w(a|s)e_{\mu}(s,a)-f\left(w(s)\right)\right] \label{Appendix-D-Lag}
\end{align}
where $e_{\mu}(s,a) = \gamma \sum_{s'}T(s'|s,a)\mu(s') - \mu(s)$, and \eqref{Appendix-D-Lag0} is derived by using the following equality:
\begin{align*}
    \sum_{s}\mu(s)(\mathcal{T}_{*}d_{w})(s))&=\sum_{s}\mu(s)\sum_{\bar{s},\bar{a}}T(s|\bar{s},\bar{a})w(\bar{s})w(\bar{a}|\bar{s})d_D(\bar{s},\bar{a})\\&=\sum_{s'}\mu(s')\sum_{s,a}T(s'|s,a)w(s)w(a|s)d_D(s,a)\\&=\sum_{s,a}w(s)w(a|s)d_D(s,a)\sum_{s'}T(s'|s,a)\mu(s')
\end{align*}
Following the assumption of OptiDICE, the strong duality holds by satisfying Slater's condition, which enables the optimization order to be switched as shown below:
\begin{align*}
    \max_{w(s) \ge 0}\min_{\mu}\mathcal{L}(w,\mu)=\min_{\mu}\max_{w \ge 0}\mathcal{L}(w,\mu)
\end{align*}
The reordering enables inner maximization over $w(s)$, whose optimal solution satisfies $\frac{\partial\mathcal{L}(w,\mu)}{\partial w(s)}=0 \ \forall s$. Optimal $w^{*}_{\mu}(s)$ can be expressed in a closed form in terms of $\mu$.
\begin{align}
w_{\mu}^{*}(s) \ = \max(0,(f')^{-1} (\mathbb{E}_{a\sim \pi_D}[w(a|s)e_{\mu}(s,a)]))\label{Appendix-D-closedform}
 \end{align}

When $w_{\mu}^{*}(s)$ is plugged into the dual function \eqref{Appendix-D-Lag}, $\mu$ loss of \textbf{Extraction} is expressed as,
\begin{align}
\label{Appendix-D-original-loss}
\min_{\mu}\mathcal{L}_{\text{ext}}(\mu) := (1-\gamma)\mathbb{E}_{s_0 \sim p_0}\left[\mu(s_0)\right]
    + \mathbb{E}_{s\sim d_D} \left[f^{*}_{0}(\mathbb{E}_{a\sim \pi_D}[w(a|s)e_{\mu}(s,a)])\right]
\end{align}

However, sample-based optimization on $\min_{\mu}\mathcal{L}_{\text{ext}}(\mu)$ is challenging due to the existence of expectations over the transition probability $T$ within $e_{\mu}(s,a)$ and the dataset policy $\pi_D$ inside the convex function $f_{0}^{*}(x)$. Using a naive single-sample estimate such as $\mathbb{E}_{(s,a,s')\sim d_D}\left[f^{*}_{0}\left(w(a|s)\left(\gamma\mu(s')-\mu(s)\right)\right)\right]$ results in significant bias.

To circumvent this bias issue, we propose a simple bias reduction technique by incorporating an additional function approximator, $A(s)$, to estimate the expectation inside $f^*_0(\cdot)$. We then decompose the $\mu$ optimization of  \eqref{Appendix-D-original-loss} into the following optimizations on $A$ and $\mu$, which share the same optimal solution of $\mu$:
\begin{subequations}
\label{Appendix-D-losses}
\begin{align}
    \min_{A} \; &\mathbb{E}_{(s,a,s')\sim d_D}\left[ \big( A(s) - w(a|s)\hat{e}_{\mu}(s,s')\big)^2 \right]\label{Appendix-D-A-loss}\\
    \min_{\mu} \; &\tilde{\mathcal{L}}_{\text{ext}}(\mu):=(1-\gamma)\mathbb{E}_{s_0 \sim p_0}\left[\mu(s_0)\right]
    + \mathbb{E}_{(s,a,s')\sim d_D}\left[(f^*_0)'(A(s))w(a|s)\hat{e}_{\mu}(s,s') \right],\label{Appendix-D-mu-loss}
\end{align}
\end{subequations}

where $\hat{e}_{\mu}(s,s') = \gamma \mu(s')-\mu(s)$.

\paragraph{\cref{proposition5.1}} 
Minimization of the objectives in \eqref{Appendix-D-original-loss} results in the same optimal $\mu^{*}$ as in \eqref{Appendix-D-losses}.

\begin{proof}
We show optimal $A^{*}(s)$ of \eqref{Appendix-D-A-loss} given $\mu$.
\begin{align*}
    A^{*}(s) &= \E_{(a,s')\sim d_D}[w(a|s)\left(\hat{e}_{\mu}(s,s')\right)]
    \\
    &= \E_{a\sim \pi_D}[w(a|s)\E_{s'\sim T}\left[\hat{e}_{\mu}(s,s')\right]]
    \\&= \E_{a\sim \pi_D}[w(a|s)e_{\mu}(s,a)], \ \forall s
\end{align*}
For simplicity in expression, we assume $\mu(s)$ is parameterized by $\theta$. We show the gradient of $\mathcal{L}_{\text{ext}}(\mu_{\theta})$ and $\tilde{\mathcal{L}}_{\text{ext}}(\mu_{\theta})$ are the same given $A^{*}(s)=\mathbb{E}_{a\sim \pi_D}[w(a|s)e_{\mu_{\theta}}(s,a)]$.
\begin{equation*}
    \frac{\partial}{\partial \theta}\mathcal{L}_{\text{ext}}(\mu_{\theta})=\frac{\partial }{\partial \theta}\tilde{\mathcal{L}}_{\text{ext}}(\mu_{\theta})
\end{equation*}
We compare the gradients of the second term of $\mathcal{L}_{\text{ext}}(\mu_{\theta})$ and $\tilde{\mathcal{L}}_{\text{ext}}(\mu_{\theta})$
\begin{align*}
    \frac{\partial}{\partial \theta}&\left(\mathbb{E}_{s\sim d_D}\left[f^{*}_{0}(\mathbb{E}_{a\sim \pi_D}[w(a|s)e_{\mu_{\theta}}(s,a)])\right]\right)
    =\mathbb{E}_{s\sim d_D} \left[(f^{*}_{0})'(A^{*}(s))\frac{\partial K_{\mu_{\theta}}(s)}{\partial \theta}\right]\\
    &=\mathbb{E}_{s\sim d_D} \left[(f^{*}_{0})'(A^{*}(s)) \mathbb{E}_{a\sim \pi_D}\left[w(a|s)\frac{\partial (\gamma \mu_{\theta}(s') - \mu_{\theta}(s))}{\partial\theta}\right] \right]\\
    &=\mathbb{E}_{(s,a)\sim d_D} \left[(f^{*}_{0})'(A^{*}(s))w(a|s)\frac{\partial (\gamma \mu_{\theta}(s') - \mu_{\theta}(s))}{\partial\theta}\right]\\
    &=\frac{\partial}{\partial \theta}\left(\mathbb{E}_{(s,a)\sim d_D}\left[(f^*_0)'(A^{*}(s))w(a|s)\hat{e}_{\mu_{\theta}}(s,s') \right]\right)
\end{align*}
where $K_{\mu_{\theta}}(s)=\mathbb{E}_{d_D}[w(a|s)\left(\gamma \mu_{\theta}(s')-\mu_{\theta}(s)\right)]$, and $\frac{\partial K_{\mu_{\theta}}(s)}{\partial \theta}=\mathbb{E}_{(a,s')}\left[w(a|s)\frac{\partial(\gamma \mu_{\theta}(s') - \mu_{\theta}(s))}{\partial \theta}\right]$.
The equivalence leads to $\frac{\partial}{\partial \theta}\tilde{\mathcal{L}}_{\text{ext}}(\mu_{\theta})=\frac{\partial }{\partial \theta}\mathcal{L}_{\text{ext}}(\mu_{\theta})$. Therefore, minimization of \eqref{Appendix-D-original-loss} results in the same optimal $\mu^{*}$ as in \eqref{Appendix-D-losses}.
\end{proof}

After the optimization on $A$ and $\mu$, the state stationary distribution correction $w(s)$, corresponding to policy correction $w(a|s)$, is obtained by substituting $A^{*}(s)$ into \eqref{Appendix-D-closedform},
\begin{align*}
w(s) = \max\left(0,(f')^{-1}(A^{*}(s))\right), \ \forall s
\end{align*}

\section{Pseudocode of CORSDICE}

\begin{algorithm}[H]
\caption{CORSDICE}
\label{alg:algorithm}
\textbf{Input}: Dataset $D,$ Initial state dataset $ p_0, \alpha$ \\
\textbf{Parameter}:~$\psi, \phi, \xi, \zeta,  \theta , \lambda$ \\
\textbf{Output}: A policy ~$\pi_\theta$
\begin{algorithmic}[]
\STATE Let $t=0$
\STATE Initialize parameters
\FOR{$t=1, 2, \ldots, N$}
\STATE Sample from $(s, a, r, c, s') \sim D$, $\red{s_0 \sim \mu_0}$
\STATE {\footnotesize {\textcolor{blue}{\# SemiDICE for penalized reward $r(s,a) - \lambda c(s,a)$}}}
\STATE Update $\nu_\psi$, $Q_{\phi}$ using \eqref{nu-loss}, \eqref{Q-loss} given $\lambda$
\STATE {\footnotesize {\textcolor{blue}{\# Estimating state distribution correction $w(s)$}}}
\STATE $w(a|s) :=\max\big(0,(f')^{-1}\big(\frac{Q_{\phi}(s,a)-\nu_{\psi}(s)}{\alpha}\big)\big)$
\STATE Update $A_{\xi}$, $\mu_\zeta$ using \eqref{u-loss}, \eqref{mu-loss} given $w(a|s)$
\STATE $w(s):=\max\left(0,(f')^{-1}\left(A_{\xi}(s)\right)\right)$
\STATE {\footnotesize {\textcolor{blue}{\# Updating the cost Lagrange multiplier}}}
\STATE Update $\lambda$ using \eqref{modified-lambda-loss} given $w(s),w(a|s)$
\STATE {\footnotesize {\textcolor{blue}{\# Policy extraction step}}}
\STATE Update $\pi_{\theta}$ using \eqref{pi-loss} given $w(a|s)$
\ENDFOR
\end{algorithmic}
\end{algorithm}

\section{D-CORSDICE}
\label{sec:appx_d_corsdice}

In Section \ref{sec:exp_diffusion}, we introduced an extended verion of CORSDICE, which utilizes diffusion~\cite{ho2020denoising,yang2020off}-based policy to compare against other baselines that adopts advanced function approximators for actor. While the most of implementation follows that of D-DICE~\cite{mao2024diffusion}, we re-state the objective function and architectural choices for the completeness.

The main contribution of D-DICE was to introduce two way of utilizing a diffusion model in DICE framework, namely \emph{guide} and \emph{select}. Given the pre-trained behavior cloning diffusion model, \emph{guide} method, as the name suggests, guides the denoising process of stochastic differential equation (SDE)-based diffusion models with the learned correction $w(a_0\mid s)$:
\begin{equation*}
    \nabla_{a_t}\log\pi_t(a_t\mid s)=\nabla_{a_t}\log\pi^D_t(a_t\mid s)+\tau\cdot\nabla_{a_t}\log\mathbb{E}_{a_0\sim\pi^D(a_0\mid a_t,s)}[w(a_0\mid s)]
\end{equation*}
where $\tau$ is a hyperparameter for scaling the guidance score, $\pi^D$ is a behavior cloned model, and $t$ is a timestep of denoising process, not MDP. Note that the compared to Eq. (6) in~\cite{mao2024diffusion}, the stationary distribution correction is replaced with the policy correction. This is because, while D-DICE stated using stationary distribution correction to guide the diffusion model, what actually used was the policy correction, as they are using semi-gradient DICE methods and semi-gradient DICE methods extract policy correction \cref{prop:solution_of_semi_dice}. However, this error does not invalidate the original D-DICE, as the relationship $\pi^*(a\mid s)=w(a\mid s)\pi^D(a\mid s)$, still holds.

In \emph{select} period, with the guided sampling, we sample multiple actions for a given state. Then, utilizing a learned $Q$ function from semi-gradient DICE method, we can choose a greedy action that maximizes $Q$-value. Two methods combined, they are called \emph{guide-then-select} method, and we adopted this method accordingly.

\subsection{Choice of \texorpdfstring{$f$}{f}-divergence}

As stated in D-DICE~\cite{mao2024diffusion}, the choice of $f$-divergence affects the stability of training diffusion model. We used Soft-$\chi^2$ divergence, introduced in OptiDICE~\cite{lee2021optidice} and defined as:
\begin{equation*}
    f_{\textrm{soft-}\chi^2}(x):=\begin{cases}
        \frac{1}{2}(x-1)^2\;&x\geq1\\
        x\log{x}-x+1\;&0\leq x< 1
    \end{cases}
\end{equation*}
D-DICE used slightly different version of $f$, where $x\geq1$ part is replaced with $(x-1)^2$. Since the difference was minor, we used the Soft-$\chi^2$ as we did in the non-diffusion CORSDICE experiment.

\subsection{Choice of Diffusion Model}

We used SDE-based diffusion model~\cite{yang2020off}, where the forward process is defined as:
\begin{equation*}
    d\mathbf{x}=f(\mathbf{x},t)dt+g(t)d\mathbf{w}
\end{equation*}
where $\mathbf{x}_0\sim p_0$ and $\mathbf{w}$ is a Brownian motion. \cite{yang2020off} demonstrated that given an arbitrary drift coefficient $f(\cdot,t):\mathbb{R}^d\to\mathbb{R}$ and a diffusion coefficient $g(\cdot):\mathbf{R}\to\mathbf{R}$, there exists an corresponding \emph{reverse} process of generating samples from the noisy data.

While the choice of drift and diffusion coefficients can be arbitrary, we adopt the Variance Preserving (VP) SDE proposed in~\cite{yang2020off} with the linear noise scheduling, given by:
\begin{align*}
    d\mathbf{x}&=-\frac{1}{2}\beta(t)\mathbf{x}dt+\sqrt{\beta(t)}d\mathbf{w}\\
    \beta(t)&=\beta_{\textrm{min}}+t(\beta_{\textrm{max}}-\beta_{\textrm{min}})
\end{align*}

For the choice of hyperparameters and network architecutres, please refer to Appendix \ref{sec:appx_dsrl}.

\section{Details of \texorpdfstring{\cref{sec:experiment_on_random_finite_mdp}}{Experiments on Random MDP}}
\subsection{Finite MDP experiment}
\label{Appendix-F-Finite}
We validate our algorithm \textbf{SemiDICE} and \textbf{Extraction} by following the experimental protocol of ~\citep{laroche2019safe, lee2020batch}. We repeat the experiment for 300 times and average the reults. In each run, an MDP with $|S|=30, |A|=4, \gamma =0.95$ is randomly generated. Initial probability $p_0(s)$ is set to be deterministic for a fixed state. We set there are four possible next states for each state-action pairs, and generate transition probability $T(s'|s,a)$ from Dirichlet distribution $[p(s_1|s,a),p(s_2|s,a),p(s_3|s,a),p(s_4|s,a)]\sim\text{Dir}(1,1,1,1)$ for every state-action pairs $(s,a)$. The reward of 1 is given to a state a single goal state that minimizes the optimal state value at the initial state; other states have zero rewards.

Assuming an offline setting, the dataset policy $\pi_D(a|s)$ is obtained by the mixture of optimal policy $\pi^{*}$ of the generated MDP and uniformly random policy $\pi_{\text{unif}}$, where $\pi_D(a|s) = 0.5 \pi^{*}(a|s)+0.5 \pi_{\text{unif}}(a|s) \ \forall s,a$. Then 30 trajectories are collected using the generated MDP and the dataset policy $\pi_D(a|s)$. Finally, we construct MLE MDP $\hat{\mathcal{M}}=\langle S, A, T_{\text{mle}}, r, p_0, \gamma\rangle$ using the offline dataset, then test the following algorithms: four DICE-based RL algorithms (\textbf{OptiDICE}, \textbf{SemiDICE}, \textbf{f-DVL}, \textbf{ODICE}), two behvaior-regularized RL algorithms (\textbf{SQL}, \textbf{XQL}), and \textbf{Extraction}, which applies the state stationary distribution extraction method to \textbf{SemiDICE}. We test the four algorithms (\textbf{OptiDICE}, \textbf{SemiDICE},\textbf{SQL}, \textbf{XQL}) over 6 different $\alpha \in \{0.0001, 0.001, 0.01, 0.1, 1.0, 10.0\}$ and two algorithms (\textbf{f-DVL}, \textbf{ODICE}) over 6 different $\beta \in \{0.1, 0.3, 0.5, 0.7, 0.9, 0.99\}$. Additional hyperparameter for \textbf{ODICE} is set to $\eta=1.0$. We note \textbf{Extraction} does not require any hyperparameters. For the choice of $f$-divergence, we adopt $\chi^2$-divergence ($f(x) = \frac{1}{2}(x-1)^2$) for SemiDICE and KL-divergence($f(x) = x\log x$) for state stationary distribution extraction. We note that state stationary distribution extraction returns the same $w(s)$ regardless of the choice of the $f$-divergence.

Offline RL policies from different algorithms are evaluated in three criteria, policy performance $\rho(\pi)$, violation of the Bellman flow constraint, and violation of the policy correction constraint. Policy performance $\rho(\pi)$ is a return collected by actually running the offline RL policy on the generated MDP. The violations are quantified using the $L_1$-norm:
\begin{align*}
    \text{viol}_{\text{B. F.}} &= \sum_s \left\lvert(1-\gamma)p_0(s)+ \gamma(\mathcal{T}_{*}d_w)(s) - (\mathcal{B}_{*}d_w)(s)\right\rvert, \\
    \text{viol}_{\text{P. C.}} &= \sum_s \Big\lvert \sum_a w(s,a)\pi_D(a|s)-1 \Big\rvert.
\end{align*}

\subsection{Offline RL performance in continuous domain}
\label{Appendix-F-continuous}
In this paper, we have described the close relationship between \textbf{SemiDICE}, semi-gradient DICE algorithms (\textbf{f-DVL}, \textbf{ODICE}), and behavior-regularized RL algorithms (\textbf{SQL}). We demonstrate that their similarities are also reflected in their practical performance, as they achieve comparable results in continuous domains. We also present the results of \textbf{OptiDICE}, which shows limited performance compared to the recent semi-gradient DICE mdthods. We evaluate the performance of the algorithms on D4RL benchmarks~\citep{fu2020d4rl}, with fixed hyperparameters as $\alpha=1$ and $\beta = 0.5$. 
\begin{center}
\begin{tabular}{ l c c c c c } 
\toprule
Task & SemiDICE & f-DVL & ODICE & SQL & OptiDICE \\
\midrule
hopper-medium & 66.2 & 63.0 & 86.1& 74.5&46.4 \\ 
walker2d-medium & 83.4 & 80.0& 84.9& 65.3& 68.1 \\ 
halfcheetah-medium & 44.7 & 47.7& 47.4& 48.7& 45.8\\
hopper-medium-replay & 73.8 & 90.7& 99.9& 95.5&20.0 \\
walker2d-medium-replay& 55.0 & 52.1& 83.6&38.2& 17.9\\
halfcheetah-medium-replay & 41.7 & 42.9& 44.0& 44.2&31.7 \\
hopper-medium-expert &110.4 & 105.8& 110.8&106.3&51.3\\
walker2d-medium-expert &109.0 & 110.1&110.8&110.2&104.0\\
halfcheetah-medium-expert &93.03 & 89.3& 93.2&39.3&59.7\\
\bottomrule
\end{tabular}
\end{center}

\section{Experiment Detail in Continuous Domain}
\label{sec:appx_dsrl}

\subsection{Dataset License}

Our agents were trained using DSRL dataset \citep{liu2024offlinesaferl}, version 0.1.0. The dataset is licensed under Creative Commons 4.0 International License (CC BY 4.0), and its supporting Python package is under the Apache License 2.0.

\subsection{Computational Cost}

To enable end-to-end training of CORSDICE and baselines, we implemented them in JAX~\cite{jax2018github}. Utilizing automatic vectorization and just-in-time (JIT) compilation, training CORSDICE for 5 seeds and 5 hyperparameter values over one million gradient update steps took approximately wall-clock time of 5,000 seconds, where the time for training single agent is approximately 200 seconds on a single RTX 3090 GPU. For D-CORSDICE, the pre-training of diffusion-based BC agent took about 2,000 seconds on a single RTX 3090 GPU.

\subsection{Evaluation Protocol}

Following DSRL, we evaluate the models with 3 different cost limits. Returns are normalized by the dataset's empirical maximum, and costs by the threshold, where a normalized cost below 1 indicates a safe agent. We increased the training seeds from 3 to 5 and gradient updates from $10^5$ steps to $10^6$ steps to ensure baseline convergence. As in DSRL, we prioritize cost-satifying, safe agents over return-maximizing, unsafe ones. We reported the highest return among safe agents, or if none exist, the return of the least-violating unsafe agents.

For experiments with advanced model, we evaluate algorithms using a single, tighter cost threshold---10 for harder Safety-Gymansium tasks, and 5 for others---averaging results over 3 seeds and 20 episodes.

\paragraph{Cost Limits} Following the process of DSRL~\cite{liu2024offlinesaferl}, we evaluated the models with 3 different cost limits. For Safety-Gymansium~\cite{ji2023safety}, cost limits of 20, 40, and 80 were used. For other two environments, BulletGym~\cite{gronauer2022bullet} and MetaDrive~\cite{li2022metadrive}, we used cost limits of 10, 20 and 40.

\paragraph{Adjustment of Cost Limit} While the objective of constrained RL~\eqref{eq:constrained_rl} assumes discounted MDP, the actual evaluation is performed with undiscounted sum of costs. This misalignment between the training objective and the evaluation protocol can be remedied by adjusting the cost limit accordingly:
\begin{equation*}
    C_{\gamma}=\tilde{C}_\textrm{lim}\cdot\frac{\left(1-\gamma^{H+1}\right)}{H}
\end{equation*}
where $H$ is the horizon of the episode. The adjusting coefficient can be derived by assuming the cost function is constant, and this adjustment is also used in DSRL~\cite{liu2024offlinesaferl}.

\subsection{Hyperparameters}

We used tanh-squashed Gaussian distribution to model the actor, and regular linear layers with ReLU activations (except for the last layer) for critic networks. Following ReBRAC~\cite{tarasov2024revisiting}, we utilized Layer Normalization~\cite{ba2016layer} in our critic networks. All networks were trained with Adam~\cite{kingma2014adam} optimizers, with the initial learning rate was set to $3e^{-4}$ and scheduled with cosine decay.

In case of D-CORSDICE, we used U-Net~\cite{ronneberger2015u} to train the score model $\epsilon_{\theta}$ where convolutional layers are replaced with regular linear layers, a common choice in diffusion-based RL~\cite{hansen2023idql,mao2024diffusion}. For sampling actions, we used DPM-solver~\cite{lu2022dpm} and their suggested configuration for sampling from conditional distribution. For training score model, AdamW~\cite{loshchilov2017decoupled} with weight decay of $1e^{-4}$ was used.

We optimized the single hyperparameter $\alpha$ for CorsDICE using random search. For D-CORSDICE, we also optimized $\alpha$ via random search, in addition to performing a grid search over two parameters: the guidance scale and the nubmer of inference actions. All baseline method were tuned using random search with the same computational budget as our method. The search ranges for common hyperparameters are summarized in \cref{tab:appx_hyperparameters}. For D-CORSDICE, we determined the search range for its specific parameter by referring to the official paper and implementation of D-DICE \citep{mao2024diffusion}.

\begin{table}[ht]
\caption{Hyperparameters and their search ranges/values for DSRL \citep{liu2024offlinesaferl} experiments. Round brackets indicate tuples, curly brackets indicate closed intervals, and square brackets indicate sets. Of these, we specifically tuned DICE $\alpha$ for CorsDICE, and $\alpha$, guidance scale, and the number of inference actions for D-CORSIDCE, informed by D-DICE \citep{mao2024diffusion}. Note that VAEs were utilized only by the baseline methods.}
\label{tab:appx_hyperparameters}
\centering
\begin{tabular}{lcccr}
\toprule
Hyperparameters & SafetyGym & BulletGym & MetaDrive \\
\midrule
Discount Factor $\gamma$ & 0.99 & 0.99 & 0.99 \\
Batch Size & 256 & 256 & 256 \\
Score Batch Size & 2048 & 2048 & 2048 \\
Soft Update $\tau$ & $5e^{-4}$ & $5e^{-4}$ & $5e^{-4}$ \\
Learning Rates & $3e^{-4}$ & $3e^{-4}$ & $3e^{-4}$ \\
\midrule
Actor hidden dims & (256, 256) & (256, 256) & (256, 256) \\
Critic hidden dims & (256, 256) & (256, 256) & (256, 256) \\
VAE hidden dims & (400, 400) & (400, 400) & (400, 400) \\
Score residual blocks & 6 & 6 & 6 \\
Score time embedding dims & 32 & 32 & 32 \\
Score conditional embedding dims & 128 & 128 & 128 \\
\midrule
DICE $\alpha$ Ranges & [0.001, 100.0] & [0.001, 100.0] & [0.001, 100.0] \\
Guidance Scale Values & \{1.0, 2.0, 4.0\} & \{1.0, 2.0, 4.0\} & \{1.0, 2.0, 4.0\} \\
Number of Inference Actions Values & \{1, 32, 64, 128\} & \{1, 32, 64, 128\} & \{1, 32, 64, 128\} \\
\bottomrule
\end{tabular}
\end{table}

\section{Additional Experiment Results}

\subsection{OptiDICE yields a state with no defined actions}
\label{sec:appx_optidice_yields_undefined_states}

\begin{figure}[ht]
\centering
\centerline{\includegraphics[width=0.5\columnwidth]{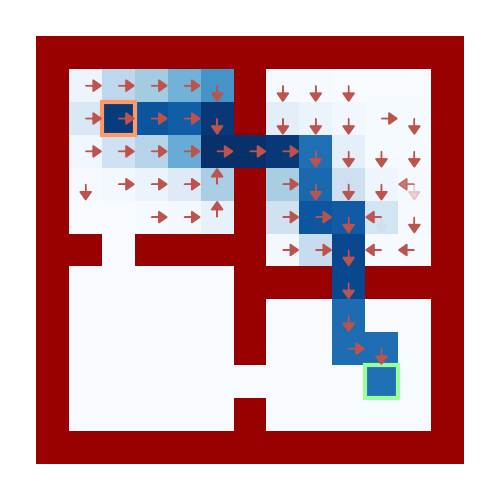}}
\caption{Visualization of state-action stationary distribution from OptiDICE. Arrow indicates the learned action with the saturation corresponding to the probability and the saturation of the cell indicates the estimated value of each state. OptiDICE yielded states with no defined actions, indicated by empty cell.}
\label{fig:optidice_fourroom}
\end{figure}

OptiDICE can yield a state with no defined action, meaning there exists a state $s$ such that $d_{\pi}^*(s,a)=0$ for all actions $a \in A$. The example of such behavior is visualized in \cref{fig:optidice_fourroom}.

\subsection{Ablation Studies on the Cost Sensitivity}
\label{sec:appx_ablation}

\begin{figure}[ht]
\vskip 0.2in
\begin{center}
\centerline{\includegraphics[width=\columnwidth]{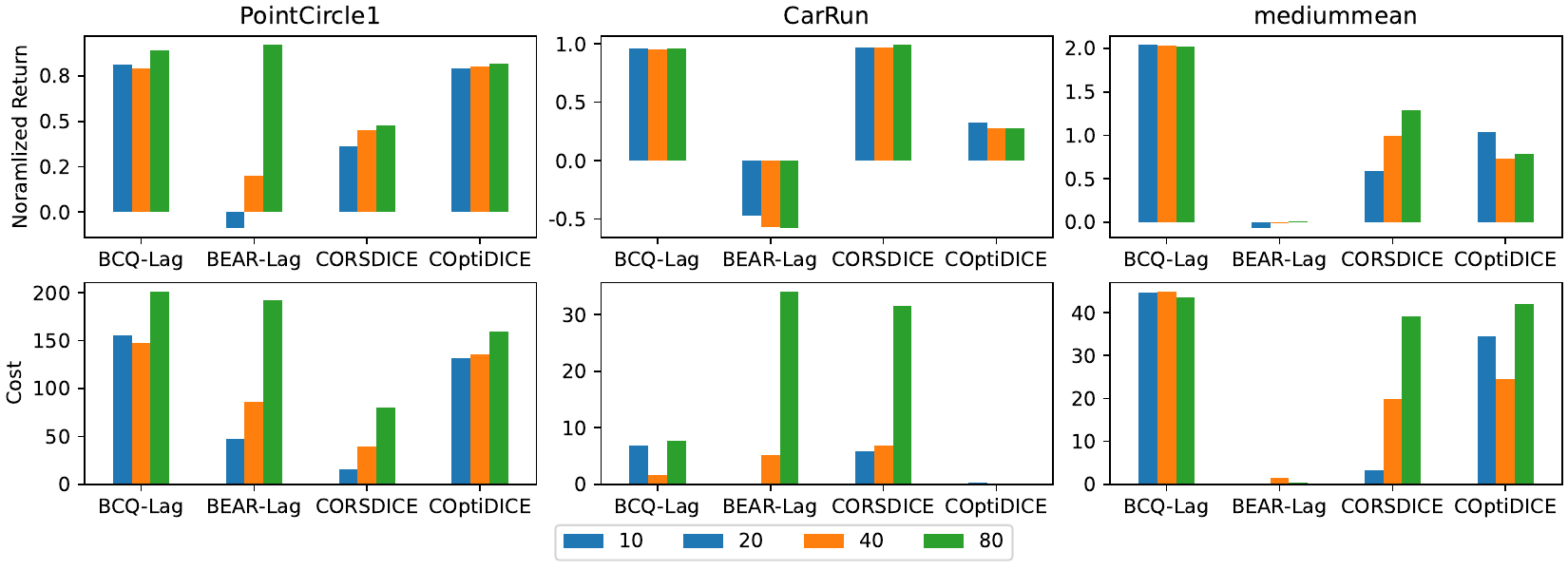}}
\caption{Ablation on the sensitivity of constrained RL algorithms on 3 different cost limits. While CORSDICE shows consistent and predictable behaviors, other baselines were inconsistent.}
\label{fig:ablation_cost_sensitivity}
\end{center}
\vskip -0.2in
\end{figure}

Effective offline constrained reinforcement learning (RL) should exhibit \emph{predictable} performance across different cost limits. Ideally, as the cost limit decreases, the algorithm should utilize fewer costs while maintaining minimal declines in return. \cref{fig:ablation_cost_sensitivity} summarizes these results. Our method, CORSDICE, demonstrated this predictable behavior, effectively using lower costs as the cost limit decreased. In contrast, other baseline methods exhibited inconsistent behavior, sometimes even incurring higher costs despite stricter cost constraints.

\subsection{Additional Experiments on Off-Policy Evaluation}
\label{sec:appx_exp_d4rl}

\begin{figure}[ht]
\begin{center}
\centerline{\includegraphics[width=0.5\columnwidth]{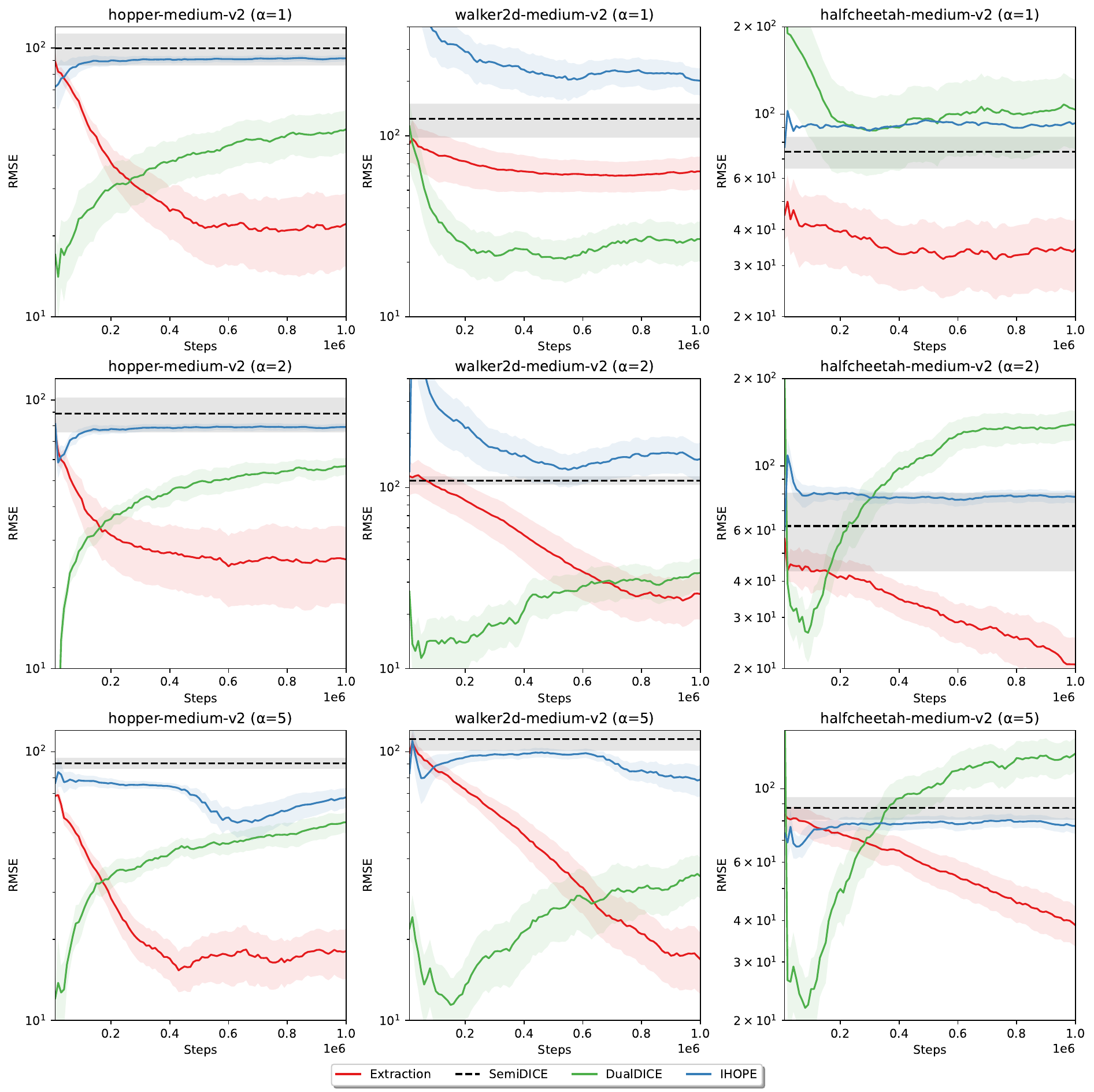}}
\caption{Root mean squared error (RMSE) of off-policy evaluation of SemiDICE policy, with different hyperparameters $\alpha$. Our extraction method is robust to the choice of $\alpha$,}
\label{fig:appx_exp_d4rl}
\end{center}
\end{figure}

We performed additional experiment to test the off-policy evaluation performance of our extraction method for different regularization strength $\alpha\in\{1, 2, 5\}$. The results are summarized in Figure~\ref{fig:appx_exp_d4rl}. Our extraction method consistently reduces the RMSE, regardless to the choice of $\alpha$.

\subsection{Learning Curves on Partial Environments}

\begin{figure}[ht]
\begin{center}
\centerline{\includegraphics[width=\columnwidth]{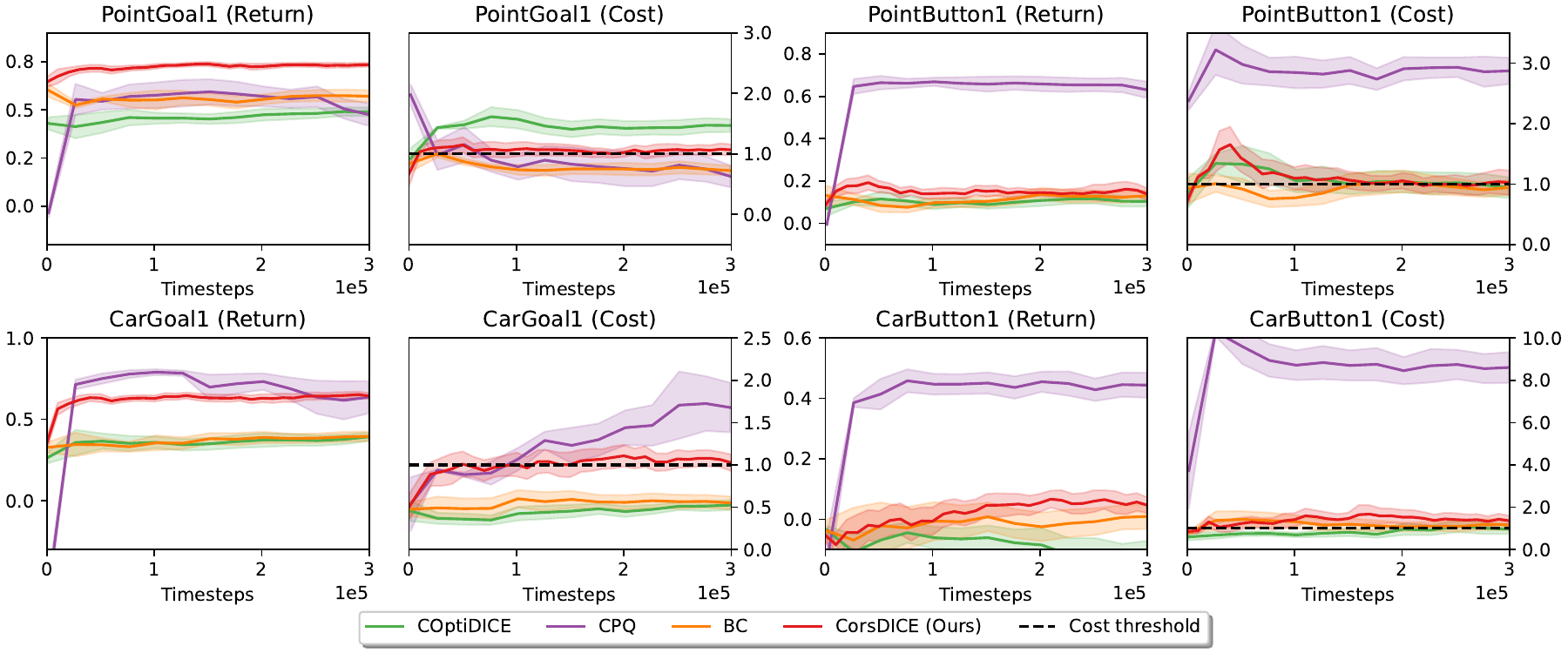}}
\caption{Early learning curves of CORSDICE and baselines on four Safety Gymnasium~\cite{liu2024offlinesaferl} tasks, cost limits set to 40. Our method, CORSDICE, shows fast and stable convergence.}
\label{fig:appx_exp_learning_curve}
\end{center}
\end{figure}

We included the learning curves of CORSDICE, including some of the baselines, to compare the convergence speed and stability on four environments, summarized in Figure \ref{fig:appx_exp_learning_curve}.

\end{document}